\documentclass[preprint]{elsarticle}
\usepackage{graphicx}
\usepackage{tabularx}
\usepackage{amssymb}
\usepackage{subfigure}
\usepackage{array}
\usepackage{lscape}
\usepackage{dcolumn,longtable,hhline,colortbl}
\usepackage[rotateright]{rotating}
\usepackage{booktabs,makecell,multirow}
\usepackage[section]{placeins}
\usepackage{amsmath}
\usepackage{enumerate}
\usepackage{color,colortbl}
\usepackage{hyperref}
\usepackage{lineno}
\usepackage[margin=1.5cm]{geometry}
\usepackage{color}
\usepackage{soul}
\usepackage{url}
\usepackage[table]{xcolor}
\usepackage{makecell}
\usepackage{amsthm}
\usepackage{multirow}
\usepackage{bm}
\usepackage{autoaligne}
\usepackage{mathtools}
\usepackage[flushleft]{threeparttable}
\usepackage{setspace}
\usepackage{appendix}
\usepackage{nomencl}
\usepackage{amsmath}
\usepackage{amssymb}
\usepackage[linesnumbered,ruled,vlined]{algorithm2e}
\usepackage{subfigure}
\usepackage{enumitem}
\usepackage{longtable}
\usepackage{multirow}
\usepackage{wrapfig}
\usepackage[flushleft]{threeparttable}
\usepackage{enumitem}
\usepackage{rotating}
\usepackage{subfigure}
\usepackage{tabularray}
\usepackage{diagbox}
\usepackage{amsmath,bm}
\usepackage[table]{xcolor}
\hypersetup{
    colorlinks=blue,
    linkcolor=blue,
    filecolor=magenta,
    urlcolor=cyan,
}
\makenomenclature

\usepackage[fontsize=11pt]{scrextend}

\renewcommand\theadalign{bc}
\renewcommand\theadfont{\bfseries}
\renewcommand\theadgape{\Gape[4pt]}
\renewcommand\cellgape{\Gape[4pt]}

\makeatletter
\hypersetup{
    colorlinks=true,
    linkcolor=blue,
    filecolor=magenta,
    urlcolor=cyan,
}

\makeatletter
\def\@xfootnote[#1]{%
  \protected@xdef\@thefnmark{#1}%
  \@footnotemark\@footnotetext}
\makeatother

\newsavebox\CBox

\newcolumntype{L}[1]{>{\raggedright\let\newline\\\arraybackslash\hspace{0pt}}m{#1}}
\newcolumntype{C}[1]{>{\centering\let\newline\\\arraybackslash\hspace{0pt}}m{#1}}
\newcolumntype{R}[1]{>{\raggedleft\let\newline\\\arraybackslash\hspace{0pt}}m{#1}}

\journal{IEEE}

\begin{document}

\begin{frontmatter}

\textbf{This work has been submitted to the IEEE for possible publication. Copyright may be transferred without notice, after which this version may no longer be accessible.}

\title{Enhancing the Performance of Neural Networks Through Causal Discovery and Integration of Domain Knowledge}

\author[address1]{Xiaoge Zhang\corref{label1}}
\ead{xiaoge.zhang@polyu.edu.hk}
\author[address2]{Xiao-Lin Wang}
\author[address3]{Fenglei Fan}
\author[address4]{Yiu-Ming Cheung}
\author[address5]{Indranil Bose}
\cortext[label1]{Corresponding author.}

\address[address1]{Department of Industrial and Systems Engineering, The Hong Kong Polytechnic University, Kowloon, Hong Kong}
\address[address2]{Business School, Sichuan University,  Chengdu 610065, China}
\address[address3]{Department of Mathematics, The Chinese University of Hong Kong, Shatin, New Territory, Hong Kong}
\address[address4]{Department of Computer Science, Hong Kong Baptist University, Kowloon Tong, Hong Kong}
\address[address5]{Department of Information Systems, Supply Chain Management \& Decision Support, NEOMA Business School, 59 rue Pierre Taittinger, Reims, 51100, France}

\begin{abstract}
Neural networks agnostic of underlying causal relationships among observed variables pose a major barrier to their deployment in high-stake decision-making contexts due to the concerns on model performance robustness and stability. In this paper, we develop a generic methodology to encode hierarchical causal structure among observed variables into a neural network for the purpose of improving its prediction performance. The proposed methodology, called causality-informed neural network (CINN), leverages three coherent steps to systematically map the structural causal knowledge into the layer-to-layer design of neural network while strictly preserving the orientation of every causal relationship. In the first step, CINN discovers causal relationships from observational data via directed acyclic graph (DAG) learning, where causal discovery is recast as a continuous optimization problem to avoid the combinatorial nature. In the second step, the discovered hierarchical causal structure among observed variables is encoded into neural network through a dedicated architecture and customized loss function. By categorizing variables as root, intermediate, and leaf nodes, the hierarchical causal DAG is translated into CINN with a one-to-one correspondence between nodes in the DAG and units in the CINN while maintaining the relative order among these nodes. Regarding the loss function, both intermediate and leaf nodes in the DAG are treated as target outputs during CINN training so as to drive co-learning of causal relationships among different types of nodes. In the final step, as multiple loss components emerge in CINN, we leverage the projection of conflicting gradients to mitigate gradient interference among the multiple learning tasks. Computational experiments across a broad spectrum of UCI datasets demonstrate substantial advantages of CINN in prediction performance over other state-of-the-art methods. In addition, we conduct an ablation study by incrementally injecting structural and quantitative causal knowledge into neural network to demonstrate their role in enhancing neural network's prediction performance. 
\end{abstract}

\begin{keyword}
Artificial intelligence\sep Causal inference\sep Causality-informed neural network\sep Computational intelligence
\end{keyword}
\end{frontmatter}


\section{Introduction}
Artificial intelligence (AI), particularly deep learning, has been commonly recognized as one of the most paramount technologies in driving the fourth industrial revolution due to its remarkable representation learning capability from massive data~\citep{lecun2015deep,gunnarsson2021deep,borchert2023extending}. Different from conventional machine learning (ML) algorithms (e.g., support vector machines, decision trees), deep learning is able to automatically extract representations from high-dimensional data in its raw form (e.g., images, voices, videos, texts) for detection or classification tasks in support of various decision-making activities. The end-to-end representation learning paradigm empowered by deep learning has produced a myriad of revolutionary innovations that even surpass human-level performance by a large margin across a broad range of tasks, such as medical diagnosis~\citep{liu2020deep}, financial market analysis~\citep{fischer2018deep,mai2019deep}, to name a few.

Despite its impressive performance, deep learning is oftentimes criticized for several critical flaws, including spurious correlations, vulnerability to small perturbations, and poor generalization~\citep{scholkopf2021toward,thuy2023explainability}, among others. These deficiencies eventually escalate and get manifested as reliability, safety, and accountability related issues on the surface in practical applications. On the one hand, the lack of these essential components in deep learning models has emerged
as an impeding roadblock to the regulatory approval of deep learning models in critical decision-making environments. On the other hand, the absence of a rigorous framework for safety assurance and risk management in deep learning models renders a low translation rate of deep learning-based methods into practical solutions in high-stake decision-making contexts. Take healthcare as an example. Only a limited number of AI-based solutions have been approved by pertinent regulatory agencies for use without human oversight, and these approved applications are predominantly limited to low-risk settings~\citep{fda_news,feng2021selective, zhang2022towards}.

While the current state-of-the-art literature has made substantial advances in addressing some of the above-mentioned issues, existing attempts to improve interpretability, robustness, and generalizability of deep learning models are, unfortunately, ad-hoc and unprincipled. In brief, post-hoc explanation methods (e.g., salience map) have been developed to associate input data with AI model's predictions through local interpretations and sensitivity analysis~\citep{simonyan2013deep,zintgraf2017visualizing,de2023explainable,chen2024interpretable}. Some approaches emphasize safeguarding AI algorithms against a certain type of adversarial attacks (e.g., data poisoning, evasion), resulting in an endless race between ``defense mechanisms” and new attacks that appear shortly after~\citep{eykholt2018robust}. Yet most theoretical developments and a large proportion of practical applications fail to tackle the most fundamental problem in deep learning, that is, how to guarantee a deep neural network learns meaningful causal relationships, rather than merely extracting correlation or association based patterns from raw data as commonly observed in the current deep learning algorithms. Since most deep learning models are agnostic to cause-and-effect relationships behind observational data, it is therefore a challenging task to guarantee the stability and robustness of prediction performance of the resultant deep learning models. In addition, a deep neural network tends to establish spurious associations from raw data, which leads to a poor generalization performance~\citep{sagawa2020investigation}. For example, \citet{ribeiro2016should} showed that a deep neural network classifier trained on a meticulously designed dataset extracts spurious correlation between snow and wolves versus Eskimo dogs (huskies) rather than meaningful features that facilitate distinguishing wolf and husky by their innate differences. Poor generalization poses a significant challenge when adopting deep learning algorithms in high-stake decision-making settings~\citep{rudin2019stop}, where input data are typically collected under dynamic environments and might not comply with the independent and identically distributed (IID) assumption foundational in deep learning. 


To tackle these issues, causal inference offers a nuanced and formal means to describe how humans comprehend the world, enabling a rigorous and sound mechanism to link causes and effects in practice~\citep{fukuyama2023dynamic,verbeke2023or,vanderschueren2023optimizing,dong2024causally}. Take computer vision as an example. Different from deep learning, humans can easily recognize an object in a scene while being less interfered by the variations in other aspects, such as background and viewing angles. If a neural network can be induced to concentrate on learning with the causal predictors in distinguishing objects, then the variations in the background and viewing angle will have a trivial effect on object recognition. The paradigm shift from correlation-based learning to causation-based learning will lead neural networks to have a stable and reliable prediction performance. In addition, reasoning with causal drivers could also extend neural network's capability beyond prediction by enabling ``what-if" analysis and interventional queries~\citep{kuzmanovic2024causal,xia2021causal,zhang2024generic}; such a capability is desirable in a broad range of industrial sectors. For example, in the telecommunication industry, in addition to identifying the users that might terminate the service, decision makers are more interested in what action to take (e.g., price discount, improving network quality) to prevent customer churn.

In fact, the ability to perform causal reasoning underscores the key difference between human intelligence and machine intelligence, and it has been recognized as a hallmark of human intelligence~\citep{pearl2019seven}.
Such capability helps to identify factors that are closely relevant to the quantity of our interest. Considering the enormous potential of causal knowledge in overcoming the innate shortcomings of traditional neural networks, this study concentrates on the confluence of causal inference and deep learning by tackling the following research question: Given causal knowledge manifested in heterogeneous forms (e.g., \emph{structural form}\textemdash in the form of a directed acyclic graph (DAG)\textemdash and \emph{relational form}\textemdash qualitative or quantitative relationships), how can we design a generic interface so that causal knowledge can be effectively and explicitly incorporated into a neural network to improve its performance? Appropriate incorporation of causal knowledge is of great significance for enhancing the performance of neural networks in several aspects. On the one hand, proper translation of causal knowledge in a structural form into neural networks largely eliminates the distraction of spurious relationships from the source that might be derived from noisy data. On the other hand, quantitative (or qualitative) causal relationships among all available (input and output) features or a subset thereof provide valuable information to guide neural network learning in the sense that they act as guardrails to prevent unexpected behaviours of neural network. From an optimization point of view, encoding causal structure and informative cause-and-effect relationships into a neural network substantially prunes the feasible region over the combination of network architecture and parameters (e.g., weights, bias), because the network is constrained to respect the underlying data generating process that is believed to produce the observational data. Exploiting causality as an inductive bias to deep learning methods is anticipated to significantly mitigate the distraction of spurious relationships and obviate to the fundamental limitations of classical deep learning models in eliciting spurious correlations. All these benefits, individually and collectively, eventually get translated into an increase in the stability and robustness of neural network performance.


In general, causal knowledge can be injected by two principal means: (1) Injecting the causal structure among observed variables into neural networks while strictly respecting the orientations associated with these causal relationships; (2) Imposing relational (quantitative or qualitative) causal relationships between certain variables as constraints in neural networks. In this paper, we propose a generic Causality-Informed Neural Network (CINN) to enforce causal structure as well as causal relationships in qualitative and quantitative forms that are either discovered from observational data or elicited from domain experts onto neural networks. The injected hierarchical causal structure ultimately gets manifested in the architecture and loss function design of the devised neural network. Different from other studies in the literature, the proposed CINN strictly adheres to the orientation of discovered causal relationships among observed variables, and offers a much more flexible interface for incorporating causal knowledge in quantitative and qualitative forms. The proposed methodology consists of three coherent steps. In the first step, we formulate  causal discovery in the form of a DAG as a continuous optimization problem following the approach developed by~\cite{zheng2018dags,zheng2020learning}, from which we derive a set of cause-and-effect relationships based on observational data. The discovered causal mechanisms are then examined by domain experts to rectify the discovered causal DAG by eliminating invalid causal links and adding substantiated causal edges. In the second step, we develop a generic methodology to fully incorporate the hierarchical causal structure among observed variables into the design of neural network architecture and loss function. In addition to accounting for the causal DAG in a hierarchical form, the proposed CINN is also capable of accommodating (partial) domain knowledge as represented by quantitative and qualitative relationships among observed variables. In the third step, since multiple loss terms appear in the developed CINN, projection of conflicting gradients (PCGrad) is adopted to mitigate gradient interference in back-propagation during the optimization of neural network parameters. Last but not the least, we design and perform an ablation study by incrementally injecting structural and relational causal knowledge into neural network to demonstrate the role of causal knowledge in boosting its prediction performance. 


Rather than restricting the application within a specific domain, this study concentrates more on developing a \emph{generic} methodology for integrating causal knowledge, and validating the methodology using a broad spectrum of datasets. The innate advantages of CINN demonstrated in these applications can be easily exploited to solve various operational research and information systems problems that are of regression nature. In summary, this paper contributes to the literature in the following perspectives. First, we develop a generic approach to explicitly mapping the hierarchical causal structure among observed variables into neural networks by establishing a one-to-one correspondence between nodes in the causal DAG and neurons in the neural network. In doing so, the causal DAG serves as an useful prior to inform the architecture design of neural network. Second, we design a dedicated loss function to minimize the total loss over the nodes in the intermediate and output layers so as to drive co-learning of causal relationships among different groups of nodes, which represents a significant departure from the existing learning paradigm. As reflected in the computational comparisons with a spectrum of baseline models, the architecture and loss function design together forms a more effective way to encode causal information of observational data into neural networks. Third, the developed CINN offers an elegant interface to capture domain expertise in two broad ways: human-guided causal discovery and domain priors on stable causal relationships. The incorporation of human knowledge largely complements the shortcomings of algorithmic causal discovery and allows multiple stakeholders (e.g., data scientist, regulators, executives) to align on the fundamental mechanics of neural network's decision-making process. Fourth, comprehensive empirical studies are conducted to examine the effect of multiple hyperparameters, including learning rate, seed value, and the weighting factor, on the CINN's performance. Computational results reveal that CINN outperforms other alternative models, demonstrating its robustness. Finally, an ablation study is performed to showcase the value of causal knowledge (causal DAG and causal relationships) in enhancing the CINN's performance incrementally. Computational results suggest that the causal graph rectified by domain knowledge and the incorporation of quantitative causal relationships play a significant role in enhancing the CINN's performance.

The rest of this paper is structured as follows. In Section~\ref{sec:literature_review}, we review the state-of-the-art approaches along the development of causal machine learning models. In Section~\ref{sec:proposed_approach}, we introduce the proposed CINN framework to enforce causal structure and causal relationships into neural network. In Section~\ref{sec:experienments}, we demonstrate the procedures of the proposed methodology, and present the computational results. In Section~\ref{sec:discussion}, we discuss the implications of proposed method for research and practice, and elaborate the limitations and future research directions. In Section~\ref{sec:conclusion}, we conclude this paper and discuss future research directions.

\section{Literature Review}\label{sec:literature_review}
This work is closely related to two streams of literature: causal discovery and causal knowledge integration into a neural network. In this section, we review the existing studies along the two directions.


\subsection{Causal Discovery}
In the past decades, there have been increasing interests along causal discovery as well as the integration of causal knowledge into neural networks~\citep{shen2020challenges,glymour2019review}. In this work, we adopt the former, and concentrate on systematically encoding hierarchical causal DAG into neural networks so that they can adhere to the orientation of each discovered causal relationship. 
In general, causal discovery with observational data is a well known NP-hard problem~\citep{silverstein2000scalable,chickering2004large} in which two important tasks emerge: \emph{structure learning}\textemdash to build a causal structure in the form of DAG~\citep{scholkopf2021toward}\textemdash and \emph{parametric learning}\textemdash to find a causal model to define the functional relationships represented by directed edges in the established causal graph~\citep{pearl2009causality}. For causal discovery purposes, Bayesian networks have been pervasively adopted in the literature because they provide a compact and sound theoretical framework for modeling causal relationships and reasoning with uncertainty over a set of random variables through a graphical model~\citep{holmes2008innovations}. Under certain conditions, the edges in a Bayesian network have causal semantics~\citep{spirtes2000causation}, thus enabling cause-and-effect analysis to some degree. In general, strategies for inducing a Bayesian network from observational data can be grouped into two categories: constraint-based methods established upon conditional independence testing and search-and-score methods. The former methods use statistical or theoretical measures to estimate whether conditional independence between some variables holds, and then infer the direction of causality between variables. Algorithms falling in this category include the well known PC algorithm~\citep{spirtes1991algorithm,spirtes2000causation} and its variants~\citep{zhang2011kernel}. The latter methods concentrate on measuring the fitness of each candidate Bayesian network structure to the observational data. Algorithms in this category search over all possible network structures and select the one that maximizes the Bayesian score (e.g., posterior score of a network given the observational data and prior knowledge) using local, greedy, or other heuristics search algorithms~\citep{heckerman2008tutorial,heckerman2013bayesian}.



Although the aforementioned methods can circumvent the enormous search space in DAG learning, they are still computationally demanding due to the combinatorial nature associated with the search space over the possible DAG structures (the number of DAGs grows exponentially with the number of observed variables) and poor search strategies. Recent research efforts have been devoted to recasting the combinatorial causal discovery problem as a continuous optimization problem that is more amenable. For example,~\citet{zheng2018dags} tackled structure learning of DAGs by converting the traditional combinatorial optimization problem into a continuous constrained one over real matrices in the linear and nonlinear contexts~\citep{zheng2018dags, zheng2020learning}. In their work, a smooth and easily differentiable function $h(W)=\text{Tr}(e^{W \circ W}) - d$ was established to enforce the acyclicity constraint in the graph, where $h(W)= 0$ if and only if $W$ corresponds to a DAG, $W \in \mathbb{R}^{d\times d}$ denotes the adjacency matrix of graph $\mathsf{G}$, and $\text{Tr}(\cdot)$ indicates the trace of a matrix. A follow-up work by~\citet{yu2019dag} proposed another acyclicity function defined as $h(W) = \text{Tr}(( I + \frac{1}{d} W \circ W)) - d$ to improve the computational efficiency. Later,~\citet{lachapelle2019gradient} extended the framework developed by~\citet{zheng2018dags} to deal with nonlinear relationships among observed variables and transformed the discrete DAG discovery problem into a pure neural network learning problem.~\citet{zhu2019causal} used actor-critic reinforcement learning algorithm to search for the DAG with the best score, where the loss function was composed of a BIC score and two penalty functions enforcing acyclicity.~\citet{pamfil2020dynotears} expanded the usage of NOTEARS to time-series data to learn dynamic Bayesian networks by leveraging the differentiable acyclicity characterization of DAG developed by~\cite{zheng2018dags}.~\citet{bello2022dagma} defined the domain of a log-determinant (log-det) function-based acyclicity characterization to be the set of $\mathsf{M}$-matrices by considering the inherent asymmetries of DAGs, and enhanced the NOTEARS's capability in detection of large cycles, gradient descent-based optimization, and computational efficiency.~\citet{gao2024causal} developed a constraint-based, non-parametric algorithm for discovering causal relations from semi-stationary time-series data, in which a finite number of causal mechanisms were repeated periodically. For a comprehensive coverage on causal discovery, refer to the recent survey paper by~\citet{vowels2022d}.

\begin{table}[!ht]\footnotesize
    \centering
    \caption{Comparison of CINN with existing related studies regarding the incorporation of causal knowledge}
    \begin{threeparttable}
    \begin{tblr}{Q[l, m, 2.9cm]|Q[c, m, 2.8cm]|Q[c, m, 3cm]|Q[c, m, 2cm]|Q[c, m, 2.2cm]|Q[c, m, 2cm]}
    \hline
    \textbf{Study} & \textbf{Form of causal knowledge} & \textbf{Means of incorporating casual knowledge} & \textbf{Preserving causal relationship orientation?}  & \textbf{Incorporating quantitative causal knowledge?} & \textbf{Incorporating structural causal knowledge?}\\
    \hline
    \citet{hu2013software}\tnote{1} & V-structure causal relationships discovered from observational data & Partial ordering constraints to enforce edge orientation among variables & \checkmark & $\times$ & \checkmark\\
    \hline
     \citet{kancheti2022matching}   & Domain prior on causal relationships  & Match the learned causal effect by neural network with desired causal effect by imposing a penalty term in the loss function & $\times$ & \checkmark & $\times$\\
    \hline
     \citet{kyono2020castle} (CASTLE)   & Implicit modeling of causal DAG   & Causal DAG learning embedded as an auxiliary task of neural network training & $\times$ & $\times$ & $\times$ \\
     \hline
     \citet{nemat2023causality} & Strength of causal effect derived from raw data &
     Causality strength are employed to weight each input variable & $\times$ & $\times$ & $\times$\\
     \hline
     \citet{russo2022causal} (CASTLE+)  & Implicit modeling of causal DAG  & Causal DAG learning embedded as an auxiliary task of neural network training & $\times$ & $\times$ & \checkmark \\
     \hline
     \citet{teshima2021incorporating}   &  Conditional independence (CI) relationship among variables & Generate new training samples to represent the CI relationship  &  $\times$ & $\times$ & $\times$ \\
     \hline
     \citet{zhai2023causality} & Causal DAG discovered from observational data & Learning with causal feature representation in graph neural network & $\times$ & $\times$ & \checkmark \\
     \hline
     \textbf{CINN} &  Causal DAG discovered from observational data, causal relationships &  Explicit mapping of hierarchical causal DAG into neural network architecture design; dedicated loss function & \checkmark & \checkmark & \checkmark \\
     \hline
    \end{tblr}
    \label{tab:comparison}
    \begin{tablenotes}
      \item[1] This study was conducted in the context of Bayesian network, not neural network.
    \end{tablenotes}
    \end{threeparttable}
\end{table}

\subsection{Causal Knowledge Integration}
In concert with the advances in causal discovery, tremendous efforts have also been dedicated to enforcing causal relationships into neural networks so as to leverage the benefits of both deep learning and cause-and-effect relationships~\citep{luo2020causal,yoon2017rnn}. For example,~\citet{kyono2020castle} proposed to incorporate Causal Structure Learning (CASTLE) as an
auxiliary task when training a neural network so as to improve its generalization performance with respect to the primary prediction task. In fact, the joint learning of causal DAG embedded in neural network training proposed by~\citet{kyono2020castle} acts as a feature selection regularizer to shrink the weights of non-causal predictors, thereby facilitating the discovery and reliance towards causal predictors. The regularization of the predictive model to conform to the causal relationships between input and target variables guarantees better out-of-sample generalization performance when handling domain shift and data perturbation. \citet{russo2022causal} proposed to improve the causal representation guarantee of CASTLE by only retaining the links present in the causal structure discovered from observational data or elicited from expert knowledge when reconstructing one feature from the remaining features.~\citet{teshima2021incorporating} developed a model-agnostic data augmentation method to leverage causal prior knowledge of conditional independence relations encoded in a causal graph for improving prediction performance of supervised ML models.~\citet{sharma2022incorporating} considered causal relationship between weather indicators and energy consumption, and extended the regular long short-term memory (LSTM) architecture using such causal relationship to augment the performance of deep learning models in the forecasting of energy demand.~\citet{wen2022causal} accounted for inter-variable causal relationships and developed a causally-aware generator with a tailored architecture to generate synthetic data that can capture target data distribution more accurately.~\citet{cui2022stable} and~\citet{zhang2021deep} formulated a stable learning paradigm in an attempt to establish common ground between causal inference and ML with the goal of enhancing model performance generalization in unseen environments.~\citet{zhai2023causality} developed a causality-based click-through rate prediction model in the graph neural networks (GNNs) framework, where causal relationships among field features were discovered and retained in GNNs via a structured representation learning approach. Recently,~\citet{berrevoets2023causal} outlined the roadmap for the development of causal deep learning framework spanning structural, parametric, and temporal dimensions to unlock its potential in solving real-world problems.

Existing methods inject causal knowledge into a neural network in an implicit manner, either via a loss function and synthetic samples encoding conditional independence or by embedding causal relationships in the latent space. Rather, the method developed in this work embraces causal knowledge in a transparent way: Each given variable in the causal graph is explicitly mapped to a specific position in the neural network in accordance with its functional role in the causal graph, and the hierarchical causal structure among observed variables is systematically translated into the design of neural network architecture, where the orientation associated with each causal relationship is preserved in the constructed neural network. In concert with the dedicated neural network architecture, CINN is devised to minimize the total loss over the nodes in the intermediate and output layers of causal graph so as to drive co-learning of causal relationships among different types of nodes. In doing so, the neural network is induced to learn correct causal relationships governing the data generation process. Moreover, aligning neural network with the hierarchical structure in causal graph opens the door for flexibly integrating domain knowledge on stable causal relationships into neural networks (see Section~\ref{sec:domain_knowledge_integration}). Hence, CINN represents a significant departure from the existing literature in the sense that it is designed with the goal of learning causal relationships among observed variables while most of deep learning models is designed solely for a single prediction task. To manifest the significance of our proposed CINN in filling these gaps in the literature, we compare it with existing related studies in Table~\ref{tab:comparison}.

\section{Methodology}\label{sec:proposed_approach}
\begin{figure}[!ht]
    \centering
    \includegraphics[scale=0.54]{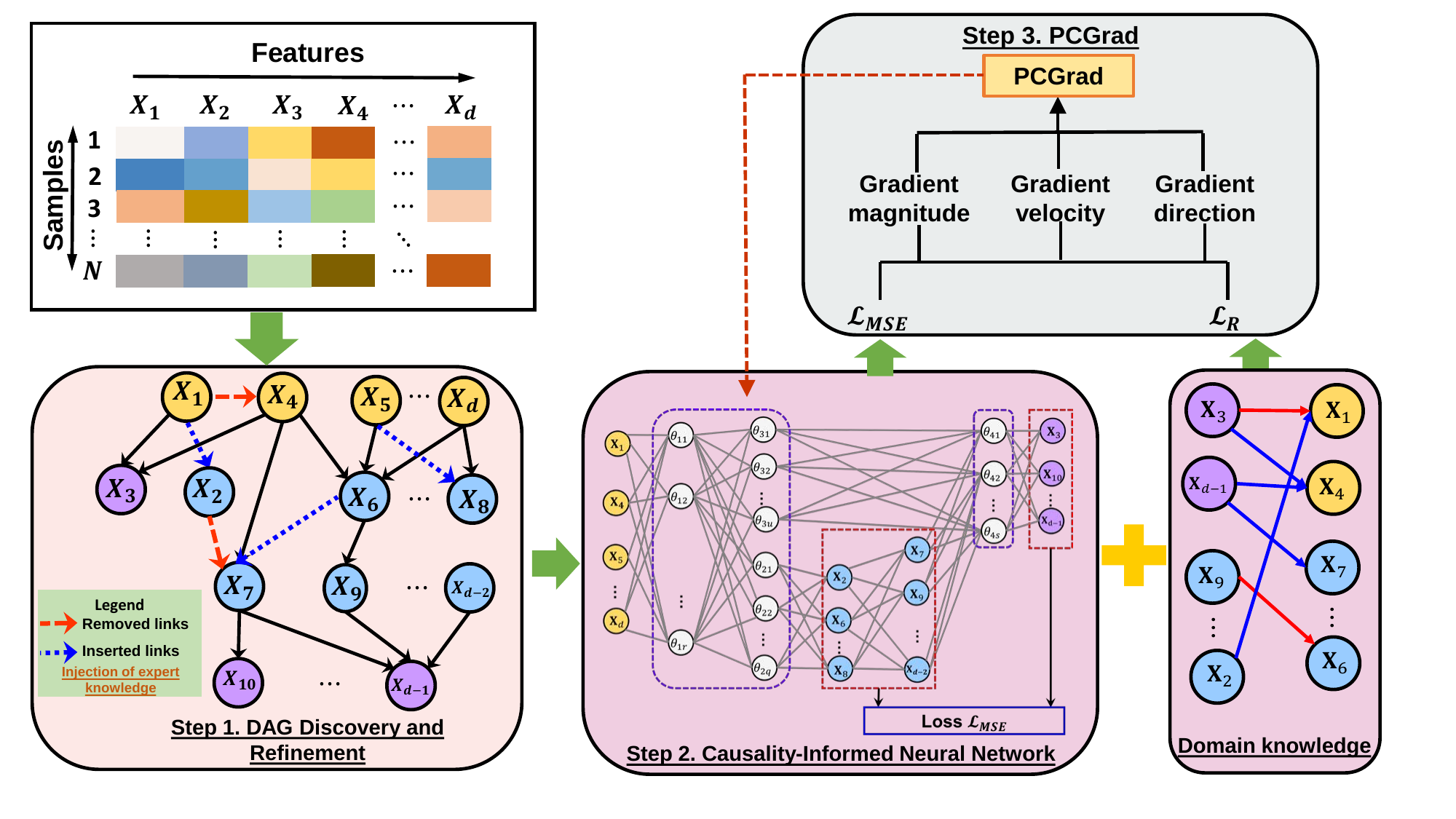}
    \caption{Flowchart of the developed methodology}
    \label{fig:Flowchart}
\end{figure}

In this section, we describe the proposed methodology for CINN in detail. The overall flowchart is illustrated in Fig.~\ref{fig:Flowchart}. As stated earlier, the developed methodology consists of three coherent steps. In the first step, causal relationships among a predefined set of variables are discovered from observational data through DAG learning. The discovered causal relationships are then examined by domain experts to eliminate any cause-and-effect relationships that are invalid, groundless, or might compromise the fairness of the algorithm, and add substantiated causal links, simultaneously. In the second step, a generic framework is developed to map the hierarchical causal structure among observed variables into the architecture and loss function design of a neural network. By aligning the neural network architecture with the causal structure among observed variables, the developed CINN is capable of accommodating quantitative or qualitative causal relationships elicited from domain experts while strictly adhering to the orientation of each discovered causal relationship. In the final step, since multiple loss components emerge in CINN, we consider loss-specific gradient features and adopt PCGrad to deconflict gradients and mitigate gradient interference towards the optimization of neural network parameters. 

\subsection{Causal Discovery and Refinement}\label{sec:causal_discovery}
Consider ${\mathsf{X}_1},{\mathsf{X}_2}, \cdots ,{\mathsf{X}_d}$ as a set of input features and $\mathsf{Y}$ as the target feature in a  regression problem. In the context of causal discovery, we observe $N$ pairs of data instances $\left(x_i, y_i \right)$, $i \in \left\{ {1,2, \cdots ,N} \right\}$, which are drawn from $\mathcal{X} = \mathcal{X}_1 \times \mathcal{X}_2 \cdots \times \mathcal{X}_d$ and $\mathcal{Y}$. In other words, ${x_i^j} \sim \mathcal{X}_j$ ($\forall i \in \left\{ {1,2, \cdots ,N} \right\}$, $\forall j \in \left\{ {1,2, \cdots ,d} \right\}$) and $y_i \sim \mathcal{Y}$, where ${x_i^j}$ indicates the $j$-th feature associated with instance $x_i$, and $\mathcal{X}_j$ $\left(j \in \left\{ {1,2, \cdots,d} \right\} \right)$ and $\mathcal{Y}$ characterize the underlying distributions in generating the $j$-th feature associated with the input and the target output, respectively.

As in~\cite{pearl2000models}, when modeling causal-and-effect relationships among random variables, we represent the causal structure over the input features ${\mathsf{X}_1},{\mathsf{X}_2}, \cdots ,{\mathsf{X}_d}$ and the target feature $\mathsf{Y}$ as a DAG. Given the observational data, the basic DAG learning problem is formulated as finding a DAG $\mathsf{G} \in \mathbb{D}$ made of vertices $\mathsf{V}=\left\{ {{\mathsf{Y}, \mathsf{X}_1},{\mathsf{X}_2}, \cdots ,{\mathsf{X}_d}} \right\}$ and edges $\mathsf{E} \subseteq \mathsf{V} \times \mathsf{V}$ so as to achieve a reasonable description on the joint distribution $P( {\mathcal{X},\mathcal{Y}})$. Provided that causal relationships among observed variables are linear, the specific relationships among the set of vertices $\mathsf{V}$ can be modelled by structural equation models, which are defined by a weighted adjacency matrix $\bm{\mathsf{W}} \in \mathbb{R}^{(d+1) \times (d+1)}$ with zeros in the diagonal.

Let $\bm{\mathsf{X}} = \left[ {\bm{\mathsf{X}}}_1, {\bm{\mathsf{X}}}_2, \cdots, {\bm{\mathsf{X}}}_d \right]$ denote the $N \times d$ input data matrix, $\bm{\mathsf{Y}}$ represent the $N$-dimensional target feature, and $\overline {\bm{\mathsf{X}}} = \left[ {\bm{\mathsf{Y}}, \bm{\mathsf{X}}} \right]$ be the $N \times \left( d + 1 \right)$ matrix containing data of all variables in the DAG. Towards causal discovery, statistical properties of the least-squares loss in scoring the DAG have been extensively investigated in the literature~\citep{van2013c0,loh2014high}: the minimizer of the least-squares loss $\frac{1}{N}\| {\bm{\mathsf{Y}} - \overline {\bm{\mathsf{X}}}\bm{\mathsf{W}}_{:, 1}} \|^2$ provably recovers a true DAG with high probability on finite samples in high dimensions. Thus, we seek to find a sparse weighted adjacency matrix $\bm{\mathsf{W}} \in \mathbb{R}^{(d+1) \times (d+1)}$ by solving the following optimization problem:
\begin{equation}
\begin{array}{l}
    \mathop {\min }\limits_{\bm{\mathsf{W}} \in \mathbb{R}^{\left(d+1 \right) \times \left(d+1 \right)}} = \frac{1}{N}\left\| {\bm{\mathsf{Y}} - \overline {\bm{\mathsf{X}}}\bm{\mathsf{W}}_{:, 1}} \right\|^2 + \lambda \left\| {\bm{\mathsf{W}}} \right\|_1, \\
    \quad \quad \text{s.t.} \quad \mathsf{H}\left( \bm{\mathsf{W}} \right) \in \mathbb{D},
\end{array}
\end{equation}
where $\bm{\mathsf{W}}_{:, 1}$ denotes the first column of $\bm{\mathsf{W}}$, $\| {\bm{\mathsf{W}}} \|_1$ denotes the $L_1$ norm of $\bm{\mathsf{W}}$, 
$\mathsf{H}( \bm{\mathsf{W}})$ defines the adjacency matrix of graph $\mathsf{G}$ (i.e., $\mathsf{H}( \bm{\mathsf{W}}) \Leftrightarrow \bm{\mathsf{W}}_{ij} \ne 0$ and zero otherwise) and $\lambda$ is a factor to weight the $L_1$ norm of $\bm{\mathsf{W}}$.

Unfortunately, the DAG constraint $\mathsf{H}\left( \bm{\mathsf{W}} \right) \in \mathbb{D}$ is discrete in nature, thus posing a significant challenge to optimization. To make this constraint more tractable, the combinatorial
acyclicity constraint is replaced with a continuous and smooth equality constraint ${\mathsf{R}(\bm{\mathsf{W}})} = 0$, where ${\mathsf{R}(\bm{\mathsf{W}})}$ measures the DAG-ness of the adjacency matrix defined by $\bm{\mathsf{W}}$. In particular, ${\mathsf{R}(\bm{\mathsf{W}})} = 0$ if and only if $\bm{\mathsf{W}}$ is acyclic (i.e., $\mathsf{H} \left(\bm{\mathsf{W}} \right) \in \mathbb{D}$). By doing so, the discrete DAG learning problem is recast as a continuous optimization problem described below~\citep{zheng2018dags}:
\begin{equation}\label{eq:no_tears}
\mathop {\min }\limits_{\bm{\mathsf{W}} \in \mathbb{R}^{\left(d+1 \right) \times \left(d+1 \right)}} = \frac{1}{N}\left\| {\bm{\mathsf{Y}} - \overline {\bm{\mathsf{X}}}\bm{\mathsf{W}}_{:, 1}} \right\|^2 + {\mathsf{R} (\bm{\mathsf{W}})} + \lambda \left\| {\bm{\mathsf{W}}} \right\|_1,
\end{equation}
where ${\mathsf{R}(\bm{\mathsf{W}})} = (\text{Tr}(e ^{\bm{\mathsf{W}} \odot {\bm{\mathsf{W}}}} - d - 1) )^2$, $\odot$ is the Hadamard product, and $e^A$ is the matrix exponential of $A$. 

By properly selecting a threshold $\tau$, we can identify the causal relationship between any two nodes as follows. If $| {\bm{\mathsf{W}}_{ij}}| \ge \tau$, then there is a causal relationship from node $\mathsf{V}_i$ to node $\mathsf{V}_j$; otherwise, no causal relationship exists. In doing so, $\bm{\mathsf{W}}$ defines a weighted adjacency matrix with a capacity to represent a broad spectrum of causal DAGs. By formulating causal DAG discovery as a continuous optimization problem over the real matrix $\bm{\mathsf{W}}$, standard numerical algorithms, such as Limited-memory Broyden–Fletcher–Goldfarb–Shanno algorithm (L-BFGS)~\citep{nocedal1999numerical}, can be leveraged to solve the non-convex optimization problem defined in \eqref{eq:no_tears}.


This step primarily serves to probe potential causal relationships among observed variables and derive an approximate structural representation over the causal relationships that is most likely to generate the observational data. Note that the discovered causal relationship between some variables might be invalid, as it might violate our common sense. To remedy such shortcoming, domain expert knowledge can be exploited to further refine the discovered graph by removing non-causal or invalid links and adding substantiated edges between certain variables. The refined causal graph is expected to better characterize practical situations.

\subsection{Causality-Informed Neural Network}\label{subsec:cinn}
\subsubsection{Node Categorization}
\begin{figure}[t!h]
    \centering
    \includegraphics[scale=0.6]{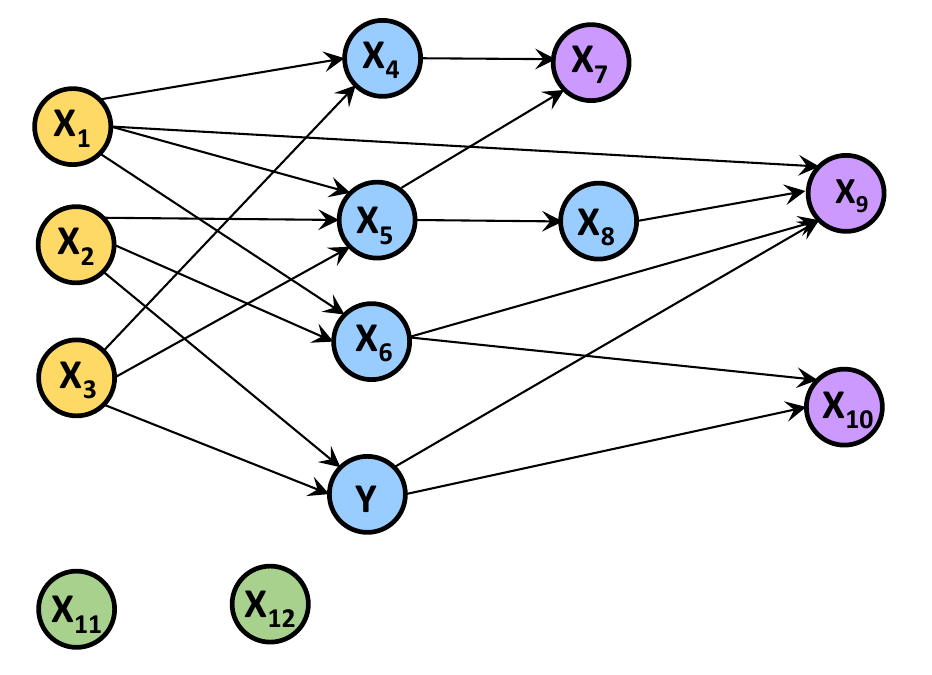}
    \caption{Demonstration of the orientation of causal relationships and node categorization. Nodes in identical color belong to the same group. Specifically, nodes in green are categorized as isolated nodes, nodes in yellow are root nodes, nodes in blue are intermediate nodes, and nodes in purple indicate leaf nodes.}
    \label{fig:node_categorization}
\end{figure}

Built upon the continuous formulation of DAG discovery, recent years have witnessed an increasing interest in leveraging the discovered causal relationships as an informative regularizer to enhance the generalization performance of neural networks. However, a noticeable flaw along the existing development of causality-aware neural networks is that they do not strictly respect the orientation of cause-and-effect relationships among observed variables, which can be either given up-front by experts or discovered from observational data. It is well known that causal relationship among variables oftentimes comes with a fixed orientation, and the direction of any causal relationship, if reversed, becomes meaningless. Unfortunately, the current paradigm of regularizing neural networks through causal graph discovery ignores the essential orientation issue. For example, \cite{kyono2020castle} proposed to regularize a neural network by masking one input feature and reconstructing it from the remaining features, where the effects as a result of the masked feature might be mistakenly used as the inputs of neural networks towards inferring the masked feature. Take the toy causal graph shown in Fig.~\ref{fig:node_categorization} as an example. In the case that $\mathsf{X}_2$ is masked out, the remaining features that include the effects of variable $\mathsf{X}_2$ (e.g., $\mathsf{X}_5$, $\mathsf{X}_6$, $\mathsf{Y}$) are used to reconstruct the masked feature $\mathsf{X}_2$. Obviously, inferring $\mathsf{X}_2$ from $\mathsf{X}_5$, $\mathsf{X}_6$, $\mathsf{Y}$ violates the orientation of the causal relationship between $\mathsf{X}_2$ and $\mathsf{X}_5$, $\mathsf{X}_6$, $\mathsf{Y}$. 

In this work, we address this issue by developing a generic framework of CINN. Unlike existing paradigms, CINN provides a rigorous and principled means to incorporate the causal structure in a hierarchical form discovered from observational data into neural networks. To elaborate on the proposed methodology, we first categorize the nodes in the discovered DAG $\mathsf{G}$ into the following four groups: 
\begin{enumerate}
    \item \textbf{Isolated nodes $\mathsf{V}_S$}: nodes without any inbound and outbound edges, such as $\mathsf{X}_{11}$ and $\mathsf{X}_{12}$ in Fig.~\ref{fig:node_categorization}.
    
    \item \textbf{Root nodes $\mathsf{V}_C$}: nodes with only outgoing edges, such as $\mathsf{X}_1$, $\mathsf{X}_2$, and $\mathsf{X}_{3}$ in Fig.~\ref{fig:node_categorization}.
    
    \item \textbf{Intermediate nodes $\mathsf{V}_B$}: nodes with both incoming and outgoing edges, such as $\mathsf{X}_4$, $\mathsf{X}_5$, $\mathsf{X}_6$, $\mathsf{X}_8$, and $\mathsf{Y}$ in Fig.~\ref{fig:node_categorization}. 
    
    \item \textbf{Leaf nodes $\mathsf{V}_O$}: nodes with only incoming edges but no successors, such as $\mathsf{X}_7$, $\mathsf{X}_9$, and $\mathsf{X}_{10}$ in  Fig.~\ref{fig:node_categorization}.
\end{enumerate}

In the case that there are multiple layers of intermediate nodes with causal relationships among themselves, the hierarchical structure of intermediate nodes can be determined by iteratively removing the current root nodes from the graph. For instance, the set of intermediate nodes for the graph shown in Fig.~\ref{fig:node_categorization} is $\mathsf{V}_B = \{ \mathsf{X}_4, \mathsf{X}_5, \mathsf{X}_6, \mathsf{Y}, \mathsf{X}_8\}$. To partition $\mathsf{V}_B$ into multiple layers of intermediate nodes, we perform the following operations iteratively:
\begin{enumerate}[label=\textbf{Iteration \arabic*:}, leftmargin=6.2em]
    \item The set of root nodes $\left\{ \mathsf{X}_1, \mathsf{X}_2, \mathsf{X}_3 \right\}$ and their associated links (e.g., $\mathsf{X}_1 \to \mathsf{X}_4, \mathsf{X}_2 \to \mathsf{X}_6, \mathsf{X}_3 \to \mathsf{X}_5$, etc.) are removed from the graph. After that, the set of nodes $\{ \mathsf{X}_4, \mathsf{X}_5, \mathsf{X}_6, \mathsf{Y} \}$ become the current root nodes. Thus, we have the first layer of intermediate node as the intersection of current root nodes and $\mathsf{V}_B$; that is, $\mathsf{V}_B^1 = \{ \mathsf{X}_4, \mathsf{X}_5, \mathsf{X}_6, \mathsf{Y} \}$.
    
    \item Next, we remove the current root nodes $\{ \mathsf{X}_4, \mathsf{X}_5, \mathsf{X}_6, \mathsf{Y} \}$ and their associated links so that $\{ \mathsf{X}_7, \mathsf{X}_8,  \allowbreak \mathsf{X}_9, \mathsf{X}_{10} \}$ become the current root nodes. Thus, the second layer of intermediate nodes is the intersection of $\{ \mathsf{X}_7, \mathsf{X}_8, \mathsf{X}_9, \mathsf{X}_{10} \}$ and $\mathsf{V}_B$; that is, $\mathsf{V}_B^2 = \{\mathsf{X}_8 \}$. 
\end{enumerate} 

The procedure continues until the number of intermediate nodes summed over all the intermediate layers $\mathsf{V}_B^1$ and $\mathsf{V}_B^2$ equals to the number of intermediate nodes $\mathsf{V}_B$ in the causal graph. In doing so, the program produces a hierarchical structure to define the type and position of each node in the DAG $\mathsf{G}$.

\subsubsection{CINN Development}
Suppose the sets of isolated nodes, root nodes, intermediate nodes, and leaf nodes in $\mathsf{G}$ are denoted by $\mathsf{V}_{S} = \{ {\mathsf{X}_{S,1}}, {\mathsf{X}_{S,2}}, \cdots {\mathsf{X}_{S,M}} \}$, $\mathsf{V}_{C} = \{ {\mathsf{X}_{C,1}}, {\mathsf{X}_{C,2}}, \cdots {\mathsf{X}_{C,T}} \}$, $\mathsf{V}_{B} = \{ {\mathsf{X}_{B,1}^{1}}, {\mathsf{X}_{B,2}^{1}}, \cdots {\mathsf{X}_{B,K[1]}^{1}; \mathsf{X}_{B,1}^{2}}, {\mathsf{X}_{B,2}^{2}}, \cdots {\mathsf{X}_{B,K[2]}^{2}; \cdots ;  \mathsf{X}_{B,1}^{R}}, \allowbreak {\mathsf{X}_{B,2}^{R}}, \cdots {\mathsf{X}_{B,K[R]}^{R}} \}$, and $\mathsf{V}_{O} = \{ {\mathsf{X}_{O,1}}, {\mathsf{X}_{O,2}}, \cdots {\mathsf{X}_{O,Z}} \}$, where subscripts $S,1; S,2; \cdots; S,M$; $C,1; C,2; \cdots; C,T$; $O,1; O,2; \cdots; \allowbreak O,Z$ are the indices of the corresponding feature $\mathsf{V}_S$, $\mathsf{V}_C$, and $\mathsf{V}_O$ in the set of vertices $\mathsf{V}$, respectively. 
Clearly, when building CINN, there is no need to account for the set of isolated nodes because they have no causal relationship with any other variables in the DAG $\mathsf{G}$.\footnote{For the case that the quantity of interest is one of the isolated nodes, there is no way to build the deep learning model because the isolated node is not causally related to any other observed variables in graph $\mathsf{G}$.} 
Regarding the set of features $\mathsf{V}_B$, $K[i]$ is a function representing the number of features in the $i$-th layer of intermediate nodes in $\mathsf{V}_B$ and $\mathsf{X}_{B,1}^i, \mathsf{X}_{B,2}^i, \cdots, \mathsf{X}_{B,K[i]}^i$ represent the corresponding features in that layer, where the superscript indicates the layer number. Naturally, we have $M + T + \sum_{i = 1}^R {K[i]}  + Z = d + 1$. 

\begin{figure}[!ht]
    \centering
    \includegraphics[scale=0.48]{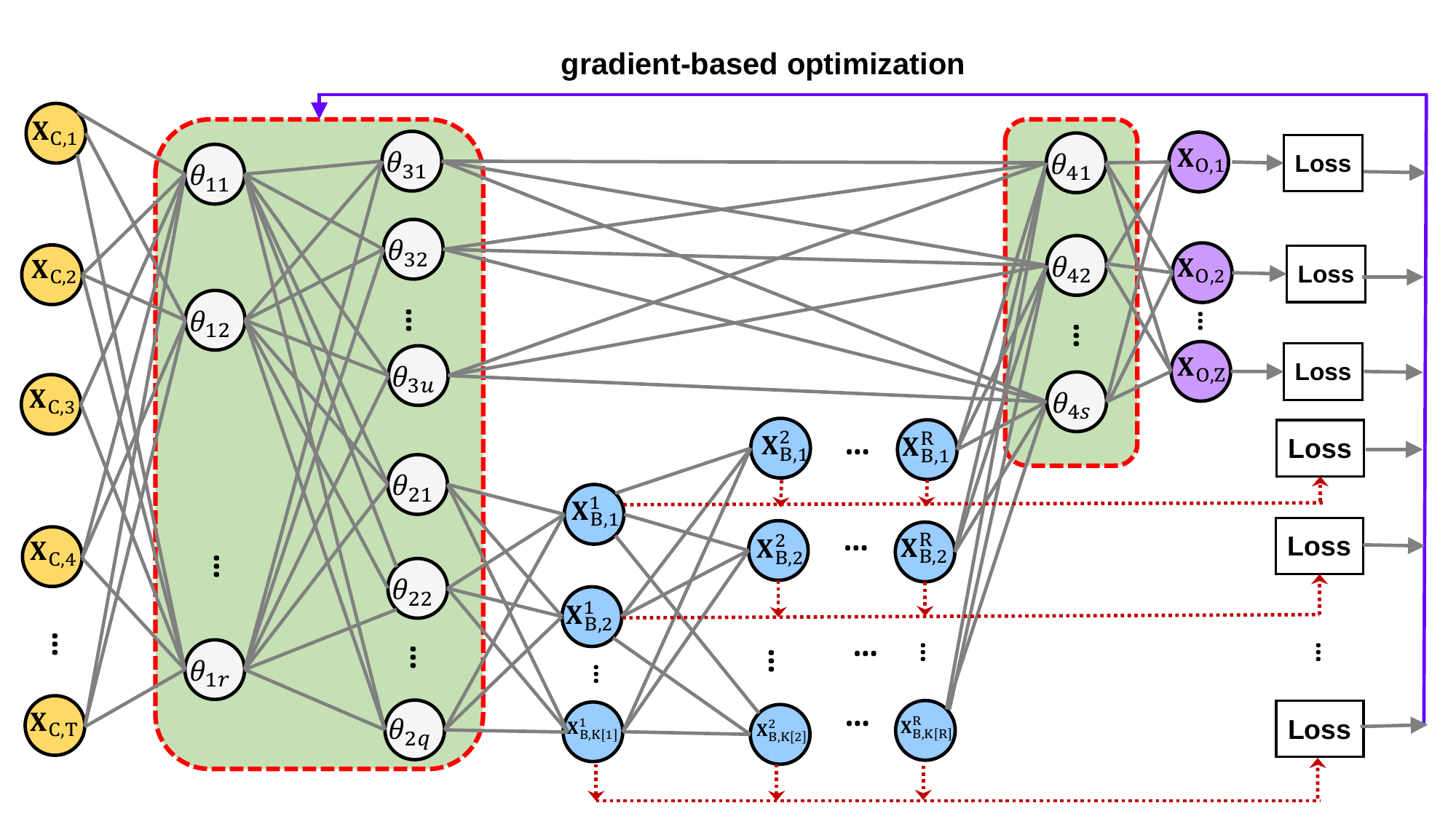}
    \caption{Proposed CINN architecture. Nodes in yellow, blue, and purple represent the sets of root nodes $\mathsf{V}_C$, intermediate nodes $\mathsf{V}_B$, and output nodes $\mathsf{V}_O$, respectively. Note that $\mathsf{V}_B$ might have multiple layers of nodes stacked together with each layer having its own set of features.}
    \label{fig:cinn_architecture}
\end{figure}

Next, we propose to align the architecture of CINN in such a way that it adheres to the causal relationships among the three node categories $\mathsf{V}_C$, $\mathsf{V}_B$, and $\mathsf{V}_O$.\footnote{The isolated nodes are discarded when creating the architecture of neural network. Thus, we refer to the number of categories as three instead of four thereafter. If one prefers to incorporate the isolated nodes in CINN, a possible way is to include them as some variables in the set of root nodes.} Fig.~\ref{fig:cinn_architecture} illustrates the CINN architecture with two hidden layers, where the causal structure in the form of DAG among the three groups of nodes $\mathsf{V}_C$, $\mathsf{V}_B$, and $\mathsf{V}_O$ is encoded into the network architecture design. More specifically, root nodes $\mathsf{V}_{C} = \{ {\mathsf{X}_{C,1}}, {\mathsf{X}_{C,2}}, \cdots {\mathsf{X}_{C,T}} \}$ act as input features of the neural network, while intermediate nodes $\mathsf{V}_{B} = \{ {\mathsf{X}_{B,1}^1}, {\mathsf{X}_{B,2}^1}, \cdots {\mathsf{X}_{B,K[1]}^1}; \cdots;  {\mathsf{X}_{B,1}^R}, {\mathsf{X}_{B,2}^R}, \cdots {\mathsf{X}_{B,K[R]}^R} \}$ play a dual role. On the one hand, they act as the outcomes of root nodes $\mathsf{V}_{C}$ or the preceding layer of intermediate nodes; on the other hand, they serve as the causes of the succeeding layer of intermediate nodes or leaf nodes $\mathsf{V}_{O} = \{ {\mathsf{X}_{O,1}}, {\mathsf{X}_{O,2}}, \cdots {\mathsf{X}_{O,Z}} \}$. Finally, leaf nodes $\mathsf{V}_{O}$ represent the outputs of the neural network, and both $\mathsf{V}_{C}$ and $\mathsf{V}_{B}$ serve as its inputs. 
It should be noted that nodes $\mathsf{V}_B$ and $\mathsf{V}_O$ share a hidden layer, denoted by $\theta_{11}, \theta_{12}, \cdots, \theta_{1r}$, in addition to their own task-specific layers of neurons as represented by $\theta_{21}, \theta_{22}, \cdots, \theta_{2q}$ and $\theta_{31}, \theta_{32}, \cdots, \theta_{3u}$, respectively. 
The purpose is to maintain the similarities and differences that coexist between the two types of nodes as observed in the original causal DAG. 
For example, as illustrated in Fig.~\ref{fig:node_categorization}, $\mathsf{X}_1$ could influence $\mathsf{X}_9$ through multiple pathways: It can affect $\mathsf{X}_9$ directly via the edge $\mathsf{X}_1 \to \mathsf{X}_9$, or indirectly through the paths $\mathsf{X}_1 \to \mathsf{X}_5 \to \mathsf{X}_8 \to \mathsf{X}_9$ and $\mathsf{X}_1 \to \mathsf{X}_6 \to \mathsf{X}_9$. This allows the effect that root nodes have on leaf nodes to be imposed via different pathways so that the effect does not necessarily be enforced through the intermediate layers.

In addition to the causality-aligned network architecture, CINN also gets embodied in the design of loss function. In a conventional neural network, there is only one target variable typically located at the output layer, and the network is trained to minimize the loss between the predicted and observed values with respect to this single target variable. 
When training CINN, however, we treat both $\mathsf{V}_B$ and $\mathsf{V}_O$ as target outputs. Our goal is to minimize the overall loss over the observations associated with $\mathsf{V}_B$ and $\mathsf{V}_O$ using the $N$ observational data so as to synergize co-learning of causal relationships among observed variables. Mathematically, the loss function of CINN is formulated as follows:
\begin{equation}\label{eq:cinn_obj}
\mathop {\min }\limits_{\bm{\theta}} \quad \mathcal{L}_{MSE} = \frac{1}{N} \sum\limits_{i = 1}^N {\left[ {\underbrace{\sum\limits_{j=1}^R
\sum\limits_{k = 1}^{K[j]} {{\left( {\overline {\bm{\mathsf{X}}}_{B,k}^{i, j} - \widehat {\bm{\mathsf{X}}}_{Bk}^{i, j}} \right)}^2}}_{\mathcal{L}_{MSE}^{B}} {\rm{ + }} \underbrace{\sum\limits_{z = 1}^Z {{{\left( {\overline{\bm{\mathsf{X}}}_{O,z}^i - \widehat {\bm{\mathsf{X}}}_{O,z}^i} \right)}^2}} }_{\mathcal{L}_{MSE}^{O}} } \right]},  
\end{equation}
where $\widehat {\bm{\mathsf{X}}}_{B,k}^{i, j}$ denotes the output with respect to neuron ${\mathsf{X}}_{B,k}^{j}$ in the $j$-th layer of intermediate nodes, and $\widehat {\bm{\mathsf{X}}}_{O,z}^{i}$ refers to the output with respect to ${\mathsf{X}}_{O,z}$ $\left( z = 1, 2, \cdots, Z \right)$ when the set of features $\mathsf{V}_C$ associated with the $i$-th observational data is fed into the neural network; ${\overline{\bm{\mathsf{X}}}}_{B,k}^{i, j}$ (resp. ${\overline{\bm{\mathsf{X}}}}_{O,z}^i$) represents the $i$-th observation associated with the feature $\mathsf{X}_{B,k}^{j}$ (resp. $\mathsf{X}_{O,z}$).

The loss function formulated in Eq.~\eqref{eq:cinn_obj} drives co-learning of causal relationships between $\mathsf{V}_C$ and $\mathsf{V}_B$, $\mathsf{V}_C$ and $\mathsf{V}_O$, as well as $\mathsf{V}_B$ and $\mathsf{V}_O$. Such an objective function would induce the trained neural network to respect the causal relationships and the orientation of causal relationships among the observed variables. By adhering to the causal structure, the trained neural network is anticipated to have a stable and augmented performance when making predictions. 

\subsubsection{Integration of Domain Knowledge}\label{sec:domain_knowledge_integration}
The developed CINN architecture illustrated in Fig.~\ref{fig:cinn_architecture} not only allows the incorporation of causal structure, but also enables encoding domain knowledge that characterizes any quantitative or qualitative relationship between any pair of cause-and-effect variables. 
For example, if a domain prior regarding the causal relationship between $\mathsf{X}_{C,t}$ $(t =1, 2, \cdots, T)$ and $\mathsf{X}_{B,k}^j$ $(k=1, 2, \cdots, K[j])$ is available, then it can be exploited to guide neural network training. Even though the exact mathematical form of such domain knowledge might be unknown, rough causal relationships in a variety of forms, such as monotonic effect, zero effect, U-shaped effect, and loose differential equation, can be treated as inequality or equality constraints when training the neural network. 
For instance, if we would like to impose a fairness constraint to ensure that race ($\mathsf{X}_{C,t}$) has no effect on the loan application ($\mathsf{X}_{B,k}^{j}$) in the learned model~\citep{kilbertus2017avoiding}, a constraint like $\text{d}{\widehat{\bm{\mathsf{X}}}_{B,k}^{j}}/\text{d}{\bm{\mathsf{X}}_{C,t}} = 0$ can be encoded into the neural network. In another case, if there is a positive causal relationship between $\mathsf{X}_{C,t}$ and $\mathsf{X}_{B,k}^{j}$, without an exact form, a soft constraint like $\text{d}{\widehat{\bm{\mathsf{X}}}_{B,k}^{j}}/\text{d}{\bm{\mathsf{X}}_{C,t}} > 0$ can be formulated. In general, the domain causal knowledge in most forms can be transformed into an equivalent differential formulation in an unified manner. Therefore, we develop a differential-based formulation to enforce the neural network to keep in compliance with the known domain prior on causal relationships:
\begin{equation}\label{eq:injection_of_quantitative_relationships}
\begin{aligned}
    \mathcal{L}_R & = \sum\limits_{i = 1}^N {\max \left\{ {0,{{\left\| {{\nabla _i}{\widehat{\bm{\mathsf{X}}}_O} \odot {\mathsf{M}_{OC}} - \delta \mathsf{g}  _{OC}^{i}} \right\|}_1} - \varepsilon } \right\}} \\
    & \quad + \sum\limits_{i = 1}^N \sum\limits_{k=1}^R {\max \left\{ {0,\left\| {{\nabla _i}{\widehat{\bm{\mathsf{X}}}_B^k} \odot {\mathsf{M}_{BC}^k} - \delta \mathsf{g}_{BC}^{i, k}} \right\|_1 - \varepsilon } \right\}} \\
    & \quad + \sum\limits_{i = 1}^N \sum\limits_{k=1}^R {\max \left\{ {0,\left\| {\frac{{{\nabla _i}\widehat{\bm{\mathsf{X}}}_O}}{{{\nabla _i}\widehat{\bm{\mathsf{X}}}_B^k}} \odot {\mathsf{M}_{OB}^k} - \delta \mathsf{g}_{OB}^{i, k}} \right\|_1 - \varepsilon } \right\}}, 
\end{aligned} 
\end{equation}
where $\mathsf{M}_{OC}$, $\mathsf{M}_{BC}^k$, and $\mathsf{M}_{OB}$ are binary matrices indicating for which variable prior causal knowledge is available. In particular, $\mathsf{M}_{BC}^k$ is a binary indicator characterizing the causal relationship between the $k$-th layer of intermediate nodes and the root nodes $\mathsf{V}_C$. 
${{\nabla _i}\widehat{\bm{\mathsf{X}}}_O}$ is a $Z \times T$ Jacobian of the $Z$-dimensional neural network prediction $\widehat{\bm{\mathsf{X}}}_O$ with respect to the $i$-th $T$-dimensional input; similarly, ${{\nabla _i}\widehat{\bm{\mathsf{X}}}_B^k}$ is a $K[k] \times T$ Jacobian of the $K[k]$-dimensional prediction $\widehat{\bm{\mathsf{X}}}_B^k$ with respect to the $i$-th $T$-dimensional input; and ${\nabla _i}\widehat{\bm{\mathsf{X}}}_O/{\nabla _i}\widehat{\bm{\mathsf{X}}}_B^{k}$ is a $Z \times K[k]$ Jacobian of the $Z$-dimensional prediction $\widehat{\bm{\mathsf{X}}}_O$ with respect to the $i$-th $K[k]$-dimensional output $\widehat{\bm{\mathsf{X}}}_B^k$; $\odot $ denotes the element-wise product. If there is prior knowledge on the causal relationship between a cause variable $i$ in $\mathsf{X}_C$ and an effect variable $j$ in $\widehat{\mathsf{X}}_O$, then $\mathsf{M}_{OC}^{ij} = 1$; otherwise, zero. The construction of $\mathsf{M}_{OB}^k$ and $\mathsf{M}_{BC}^k$ follows a similar approach. $\delta \mathsf{g}_{OC}^i$, $\delta \mathsf{g}_{BC}^{i, k}$, and $\delta \mathsf{g}_{OB}^{i, k}$ denote matrix derivatives of available domain priors with a size of $Z \times T$, $K[k] \times T$, and $Z \times K[k]$ with respect to $\mathsf{X}_C$, $\mathsf{X}_C$, and $\widehat {\mathsf{X}}_B$, respectively. $\varepsilon$ is a hyperparameter to allow a margin of error when regularizing the neural network with prior causal knowledge. 

By incorporating domain knowledge into the neural network, we formulate a composite loss function to guide the training of CINN as informed by the rectified causal structure and the prior causal knowledge:
\begin{equation}\label{eq:overall_loss}
    \mathcal{L} = \mathcal{L}_{MSE} + \gamma \mathcal{L}_R,
\end{equation}
where $\gamma$ denotes the weight associated with the domain prior term $\mathcal{L}_R$. Increasing (resp. lowering) the value of $\gamma$ strengthens (resp. diminishes) the regularization effect of prior domain knowledge. Devising the architecture and loss function in such a manner brings substantial benefits to neural network learning. 

\begin{enumerate}
    \item[i)] Incorporating the hierarchical causal structure can diminish the distraction of spurious relationships derived from noisy data to a neural network~\citep{kyono2020castle,russo2022causal,fan2022debiasing,hu2022improving}. Let us revisit the causal graph shown in Fig.~\ref{fig:node_categorization}. Suppose there is a strong spurious correlation between $\mathsf{X}_{9}$ and $\mathsf{X}_{10}$, and $\mathsf{X}_{9}$ is the quantity of our interest. If $\mathsf{X}_{10}$ is used as an input feature, then deep learning model tends to rely more on $\mathsf{X}_{10}$ to predict $\mathsf{X}_{9}$ rather than other upstream causal factors (e.g., $\mathsf{X}_{1}$, $\mathsf{X}_{5}$, $\mathsf{X}_{8}$, $\mathsf{Y}$), because $\mathsf{X}_{10}$ helps to reduce the loss the most. Encoding the causal structure into a neural network prevents the learning of spurious relationships as the network architecture is aligned with the structural causal relationships as defined in the discovered causal graph. The informative network architecture serves as a good starting point towards stable and meaningful representation learning. 
    
    \item[ii)] Different from existing studies, the developed neural network architecture strictly respects the orientation of causal relationships that are believed to govern the observational data. The orientated causal relationships encoded into neural network can prune the search space over possible combinations of network architecture and parameters, thus providing valuable information to effectively guide neural network training. Such an informative architecture design makes neural network more parsimonious by adhering to the established causal structure among the observed variables. Moreover, the proposed CINN provides straightforward interfaces for integrating domain knowledge into neural networks. On the one hand, domain knowledge can be leveraged to refine the hierarchical causal structure discovered from observational data by pruning the discovered causal graph $\mathsf{G}$, such as removing unreasonable edges, adding causal relationships between certain variables, among others. On the other hand, CINN allows the incorporation of causal relationships among many pairs of observed variables as specified by domain experts. Specifically, it not only allows the incorporation of causal relationships between input features ${\mathsf{X}_1},{\mathsf{X}_2}, \cdots ,{\mathsf{X}_d}$ and target output $\mathsf{Y}$, as some existing studies have done, but also enables encoding causal relationship between two variables in input features ${\mathsf{X}_1},{\mathsf{X}_2}, \cdots ,{\mathsf{X}_d}$, if any. 
    
    \item[iii)] The causal structure discovered from observational data is amendable to available domain priors on causal relationships, thereby facilitating the combination of well established expert knowledge and observational data. In fact, combining observational data and domain knowledge has been widely adopted in the existing literature to gain a better structural representation of relationships among random variables, such as Bayesian network learning~\citep{heckerman1997bayesian,hu2013software,von2021informed}.
\end{enumerate}

\subsection{Projection of Conflicting Gradients for CINN}\label{sec:pcgrad}
As shown in Eq.~(\ref{eq:overall_loss}), the loss function for CINN has two terms, namely, $\mathcal{L}_{MSE}$ and $\mathcal{L}_R$. The former consists of two individual components $\mathcal{L}_{MSE}^B$ and $\mathcal{L}_{MSE}^O$ that quantify the losses specific to the predictions associated with $\mathsf{V}_B$ and $\mathsf{V}_O$, respectively, while the latter measures the degree of neural network in violating the imposed domain knowledge on causal relationships. Note that every single component $\mathcal{L}_{MSE}^B$ and $\mathcal{L}_{MSE}^O$ might consist of a series of loss items; see Fig.~\ref{fig:cinn_architecture} for more information. For example, $\mathcal{L}_{MSE}^B$ might be composed of 10 intermediate outputs.\footnote{The losses can also be fed into PCGrad in a granular way by treating the loss associated with each target output (e.g., nodes $\mathbf{x}_{B,1}^1, \mathbf{x}_{B,2}^2, \cdots, \mathbf{x}_{B,K[R]}^R$, $\mathbf{x}_{O,1}, \mathbf{x}_{O,2}, \cdots, \mathbf{x}_{O,Z}$ in Fig.~\ref{fig:cinn_architecture}) as an individual learning task.}

Obviously, it is a challenging task to train CINN to master multiple tasks simultaneously, because multiple learning objectives are highly involved~\citep{vandenhende2021multi}. In the back-propagated optimization of neural network weights via gradient descent-based optimizers, the three loss terms $\mathcal{L}_{MSE}^B$, $\mathcal{L}_{MSE}^O$, and $\mathcal{L}_{R}$ are typically averaged by a weighted sum. In the case that the three loss terms have significant differences in magnitude, the average operation can make some loss component dominate the others during the learning process. Optimizing towards the dominated task leads to performance degradation and sacrifice of other learning tasks. In addition to imbalance in gradient magnitude, the gradients of different learning tasks might be conflicting along the direction of descent with one another, which is detrimental to the optimization of neural network parameters. As a result, the optimizer struggles back and forth when optimizing the weights of neural network.

In this work, we tackle this problem from a multi-task learning perspective, where each loss component is treated as an individual learning task. To mitigate gradient interference among different learning tasks, we adopt the PCGrad approach to eliminate conflicting elements from the set of gradients~\citep{yu2020gradient}. The key idea is to project the gradient of a task onto the norm plane of any other task if there is a conflict between their gradients. To demonstrate the idea of PCGrad, we take the gradients associated with $\mathcal{L}_{MSE}^B$ and $\mathcal{L}_{R}$ as an example. PCGrad first checks whether there is any conflict between $\bm{\Delta}_{\mathcal{L}_{MSE}^B}$ and $\bm{\Delta}_{\mathcal{L}_{R}}$ using the cosine similarity metric defined below:
\begin{equation}\label{eq:similarity}
\omega \left( {{\bm{\Delta}_{\mathcal{L}_{MSE}^B}},{\bm{\Delta}_{\mathcal{L}_{R}}}} \right) = \frac{{{\bm{\Delta}_{\mathcal{L}_{MSE}^B}} \bullet {\bm{\Delta}_{\mathcal{L}_{{R}}}}}}{{\left\| {{\bm{\Delta}_{\mathcal{L}_{MSE}^B}}} \right\|_2\left\| {{\bm{\Delta}_{\mathcal{L}_{{R}}}}} \right\|_2}},
\end{equation}
where $\|\cdot \|_2$ denotes the $L_2$ norm of a vector. The metric produces a value within the range $[- 1,1]$, where $-1$ denotes exactly the opposite direction, 1 means exactly the same direction, and 0 indicates orthogonality or decorrelation. 

\begin{figure}[!ht]
    \centering
    \includegraphics[scale=0.56]{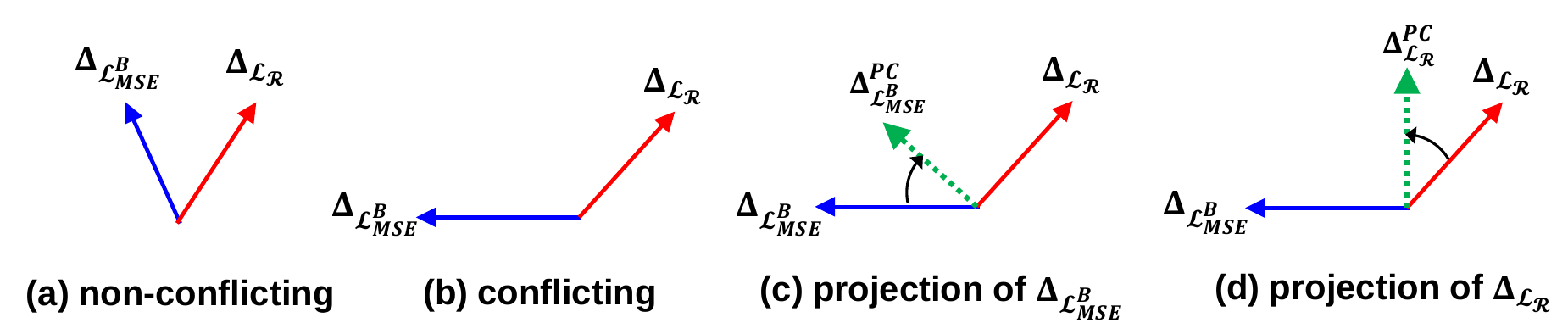}
    \caption{Demonstration of PCGrad. (a) There is no conflict between $\bm{\Delta}_{\mathcal{L}_{MSE}^B}$ and $\bm{\Delta}_{\mathcal{L}_{R}}$. (b) There is a high conflict between $\bm{\Delta}_{\mathcal{L}_{MSE}^B}$ and $\bm{\Delta}_{\mathcal{L}_{R}}$. (c) PCGrad projects the gradient $\bm{\Delta}_{\mathcal{L}_{MSE}^B}$ onto the norm vector of the gradient $\bm{\Delta}_{\mathcal{L}_{\mathcal{R}}}$. (d) PCGrad projects the gradient $\bm{\Delta}_{\mathcal{L}_{\mathcal{R}}}$ onto the norm vector of the gradient $\bm{\Delta}_{\mathcal{L}_{MSE}^B}$.}
    \label{fig:pcgrad}
\end{figure}

If $\omega ( {{\bm{\Delta}_{\mathcal{L}_{MSE}^B}},{\bm{\Delta}_{\mathcal{L}_{R}}}} ) < 0$, then PCGrad projects $\bm{\Delta}_{\mathcal{L}_{MSE}^B}$ to the norm plane of $\bm{\Delta}_{\mathcal{L}_{R}}$ or the other way around; otherwise, $\bm{\Delta}_{\mathcal{L}_{MSE}^B}$ and $\bm{\Delta}_{\mathcal{L}_{R}}$ remain the same (see Fig.~\ref{fig:pcgrad} for illustration). For demonstration, suppose we project $\bm{\Delta}_{\mathcal{L}_{MSE}^B}$ to the norm plane of $\bm{\Delta}_{\mathcal{L}_{\mathcal{R}}}$. Then the gradient of $\bm{\Delta}_{\mathcal{L}_{MSE}^B}$ after the projection operation becomes 
\begin{equation}
\bm{\Delta}_{\mathcal{L}_{MSE}^B}^{PC} = \bm{\Delta}_{\mathcal{L}_{MSE}^B} - \frac{{{\bm{\Delta}_{\mathcal{L}_{MSE}^B}} \bullet {\bm{\Delta}_{\mathcal{L}_{\mathcal{R}}}}}}{{{{\left\| {{\bm{\Delta}_{\mathcal{L}_{\mathcal{R}}}}} \right\|}^2}}}{\bm{\Delta}_{\mathcal{L}_{\mathcal{R}}}},
\end{equation}
where $\bm{\Delta}_{\mathcal{L}_{MSE}^B}^{PC}$ denotes the gradient of $\bm{\Delta}_{\mathcal{L}_{MSE}^B}$ after projection.

\begin{algorithm}[!ht]
\SetAlgoLined
\SetNoFillComment
\SetKwFor{For}{for (}{) $\lbrace$}{$\rbrace$}
\caption{Projecting conflicting gradients in CINN}\label{alg:flowchart}
\KwData{neural network weights $\bm{\theta}$}

$\bm{\Delta}_i$ $\leftarrow $ ${\nabla _{\bm{\theta}} }{\mathcal{L}_i}\left( {\bm{\theta}}  \right),\;\;\forall i = 1,2,3$ \tcc{$\mathcal{L}_i$ denotes the $i$-th loss term. Note $\mathcal{L}_1 = \mathcal{L}_{MSE}^B$;$\mathcal{L}_2 = \mathcal{L}_{MSE}^O$;$\mathcal{L}_3 = \mathcal{L}_{R}$.}

$\bm{\Delta} _i^{PC} \leftarrow {\bm{\Delta} _i}, \; \; \forall i$\

\For{$i = 1;\ i <= 3;\ i = i + 1$}{
    \For{$j\mathop  \sim \limits^{\mathrm{ly}} \left[ {1,2, 3} \right],\;$ $\mathrm{where}$ $m \ne i$}{
    
      \If{$\omega \left( {{\bm{\Delta} _i^{PC}},{\bm{\Delta} _j}} \right) < 0$}{
        Set $\bm{\Delta} _i^{PC} = \bm{\Delta} _i^{PC} - \frac{{{\bm{\Delta} _i^{PC}} \bullet {\bm{\Delta} _j}}}{{{{\left\| {{\bm{\Delta} _j}} \right\|}^2}}}{\bm{\Delta} _j}$\ \tcc{Subtract the projection of $\bm{\Delta} _i^{PC}$ onto $\bm{\Delta}_j$}
      }
    }
}
\Return $\bm{\Delta} \bm{\theta}  = \sum\limits_{i = 1}^\mathcal{3} {\bm{\Delta} _i^{PC}} $ 
\end{algorithm}

Algorithm~\ref{alg:flowchart} describes the implementation of PCGrad in CINN. PCGrad repeats the projection operation for all the learning tasks in a random order. In essence, the gradient projection operation in Fig.~\ref{fig:pcgrad} amounts to removing the conflicting element from the gradients, thus mitigating destructive gradient interferences among different learning tasks. The introduction of PCGrad in CINN frees us from the tuning of weights associated with each loss term. In addition, since the gradient projection operation accounts for gradient information (e.g., magnitude, direction) associated with all the loss components in a holistic manner, it significantly mitigates the conflicts among different learning tasks and produces a set of gradients with a minimal gradient interference, thus leading to a stable convergence in the optimization of neural network parameters.

\section{Computational Experiments}\label{sec:experienments}
In this section, we comprehensively examine the prediction performance of the developed CINN methodology on a broad spectrum of publicly available datasets. The performance of CINN is compared against a wide range of state-of-the-art models in the literature. The robustness of CINN with respect to several hyperparameters (e.g., learning rate, seed value for neural network initialization, weighting factor $\gamma$) is examined using a series of computational experiments. In addition, we also conduct an ablation study to demonstrate the value of leveraging causal knowledge in refining the prediction performance of CINN.


\subsection{Datasets}
We consider five datasets that are publicly available in the UCI Machine Learning Repository~\citep{asuncion2007uci}, including Boston Housing (BH), Wine Quality (WQ), Facebook Metrics (FB), Bioconcentration (BC), and Community Crime (CM), to demonstrate how CINN generates better predictions compared to other prevailing models. These datasets represent a broad range of application domains with tasks varying from house price prediction (BH), crime rate prediction (CM) to the forecast on the number of interactions to a post in Facebook (FB) as well as wine quality prediction (WQ). These datasets also vary in terms of the number of variables, the number of edges among variables, causal relationships among observed variables, available data for model development, the distribution of target variable values, and represent a diverse spectrum of sectors. Importantly, causal relationships are ubiquitously present in the observed data across the five considered datasets. For example, in the case of BH, per capita crime rate by town (CRIM) and the average number of rooms per dwelling (RM) are causally associated with the median value of owner-occupied homes (MEDV). In fact, these factors make the five datasets a common choice for causal discovery and development of causally-aware ML models \citep[see, e.g.,][]{kyono2020castle, russo2022causal, zhang2011kernel}.

By demonstrating the CINN's performance on these different prediction tasks, our goal is to showcase the generalization of CINN's prediction performance in different business contexts. Table~\ref{tab:dataset_description} briefly describes the basic information of each dataset used for performance validation and comparison, such as the features and the target variable. Note that after an ML model is trained, we verify its prediction performance only regarding the target variable (referred to as the \emph{primary prediction task} throughout the rest of the paper). 

\begin{table}[!ht]\small
    \centering
    \caption{Descriptions of datasets used in the computational experiments}
    \begin{threeparttable}
    \begin{tabular}{lC{2em}R{3em}L{20em}L{17em}}
    \toprule
      dataset   &  $d+1$ & $N$ & Example features & Target variable \\
      \midrule
      BH     & 14 & 506  & Crime rate by town, average number of rooms & Median value of owner-occupied homes \\
      WQ     & 12 & 4,898  & Fixed acidity, volatile acidity, residual sugar & Wine quality\\

      FB     & 19 & 500  & Category, page total likes, post month, post weekday, post hour, type & Total interactions (comments + likes + shares)\\
      BC     & 14 & 779  & Number of heavy atoms, average valence connectivity & Bioconcentration factor in log units\\


      CM     & 128 & 1994 & Population for community, median household income &  Number of violent crimes per 100K population\\
      \bottomrule
    \end{tabular}
    \label{tab:dataset_description}
    \begin{tablenotes}
      \item Note: $N$ is the number of observations, and $d+1$ is the number of features including the target variable $\mathsf{Y}$.
    \end{tablenotes}
    \end{threeparttable}
\end{table}

\subsection{Baselines}
As we are developing a new way to encode cause-and-effect relationships into neural networks, we restrict our comparison to state-of-the-art methods that exploit causality as a mechanism to regularize neural network for improving its prediction performance. In addition, we also compare CINN with other prevailing regularizers commonly adopted to enhance the prediction performance of neural networks. To this end, we benchmark the proposed CINN against the following methods:
\begin{itemize}[leftmargin=*]
    \item \textbf{Early stopping}~\citep{goodfellow2016deep}: Early stopping is a prevailing regularization method used in deep neural networks to halt model training when parameter updates no longer yield an improvement on the validation set. In essence, when this is the case (after a number of iterations), we terminate model training and use the best-performing parameters obtained so far as the parameters of a neural network. Early stopping is used as a baseline in the rest of the paper. 

    \item \textbf{$L_1$ norm}: $L_1$ regularization (also referred to as Lasso regularization) employs the sum of the absolute values of the weight parameters in neural network as a constitutional part of the loss function to reduce model complexity. In doing so, $L_1$ norm promotes sparsity by forcing some weights to zero, thus preventing the model from overfitting.

    \item \textbf{$L_2$ norm}: $L_2$ regularization adds the sum of squared magnitude of model parameters as a penalty term to the loss function. As model complexity increases, the penalty term penalizes larger weight values to regularize neural network so that the trained model generalizes well to unseen data.

    \item \textbf{Dropout (DO)}~\citep{srivastava2014dropout}: DO randomly drops out hidden and visible units in a neural network to prevent neurons from co-adapting excessively. Through random dropout, DO breaks up co-adaptions by making the presence of any particular hidden unit unreliable. In doing so, DO prevents neural network from overfitting and thus increases its generalization to unseen data.

    \item \textbf{Batch normalization (BN)}~\citep{ioffe2015batch}: BN is a mechanism to accelerate the training of deep neural network through a normalization step that fixes the means and variances of each layer's inputs over every mini-batch. The normalization operation stabilizes the training of neural network and permits us to use much higher learning rates when training neural network.

    \item \textbf{Input noise}~\citep{krizhevsky2012imagenet,lecun2015deep}: Adding a small amount of random noise to inputs helps prevent neural network from memorizing all the training examples, particularly in the case of small dataset. The randomly injected noise reduces generalization error and increases robustness. 

    \item \textbf{Mixup}~\citep{zhangmixup}: Mixup is a data-agnostic augmentation technique to construct virtual training instances from a pair of samples randomly selected from the training dataset. The mechanism behind mixup is to regularize a neural network by favoring simple linear behavior in-between training examples.   
    
    \item \textbf{CASTLE}: As developed by~\cite{kyono2020castle}, CASTLE incorporates causal DAG discovery as an auxiliary task to regularize neural network weights such that the weights of those non-causal predictors are shrunk. Fig.~\ref{fig:CASTLE} gives a schematic description on CASTLE. As can be observed, in the auxiliary task of causal DAG learning, CASTLE reconstructs the masked feature (e.g., $\bm{\mathsf{X}}_k, \; k = 1,2, \cdots ,d$) from the remaining features ($\bm{\mathsf{Y}}, \bm{\mathsf{X}}_1, \cdots, \bm{\mathsf{X}}_{k-1}, \bm{\mathsf{X}}_{k+1}, \cdots, \bm{\mathsf{X}}_d, k = 1,2, \cdots ,d$) subject to the DAG constraint ${\mathsf{R}(\bm{\mathsf{W}})} = 0$. In doing so, CASTLE imposes the $d$ subnetworks of learning causal DAG as a joint task when training the neural network. Hence, CASTLE exploits causal DAG discovery to implicitly incorporate a causal graph into a neural network via a tailored loss function consisting of DAG loss $\mathsf{R}(\bm{\mathsf{W}})$, predictor reconstruction loss $f_k (\overline {\bm{\mathsf{X}}})$, and prediction loss $\| {{\bm{\mathsf{Y}}} - \overline {\bm{\mathsf{X}}}\bm{\mathsf{W}}} \|$.

    \begin{figure}[!ht]
        \centering
        \includegraphics[scale=0.5]{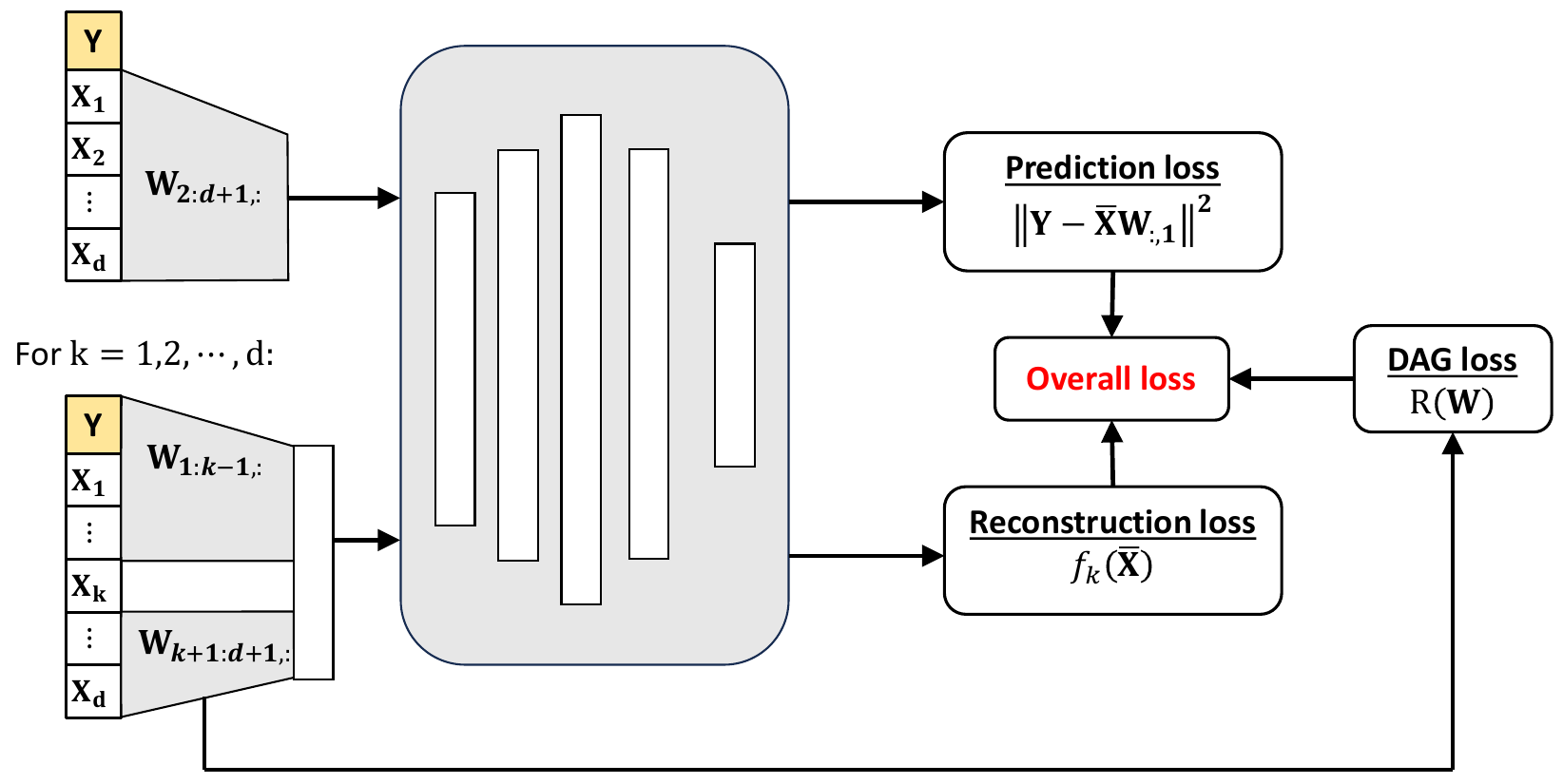}
        \caption{Graphical illustration of CASTLE developed by~\cite{kyono2020castle}}
        \label{fig:CASTLE}
    \end{figure}
    
    \item \textbf{CASTLE+}: While CASTLE reconstructs each masked predictor from the remaining predictors and $\bm{\mathsf{Y}}$, CASTLE+~\citep{russo2022causal} attempts to improve the performance of CASTLE by enforcing a causal DAG $\mathsf{G}^c (\mathsf{E}^c, \mathsf{V}^c)$ from domain experts so that the learned model is guaranteed to abide by the causal relationships defined in $\mathsf{G}^c$. Considering a given edge from $\mathsf{X}_i$ to $\mathsf{X}_j$, if the edge does not exist in the causal graph $\mathsf{G}^c$ approved by domain experts, then CASTLE+ masks the weight of the input layer as zero in the $d$ subnetworks for predictor reconstruction; otherwise, masking is not applied in the $d$ subnetworks.

\end{itemize}

In this work, we benchmark CINN against the above-described methods in no particular order. Note that BN and DO are applied after every dense layer, and they are active only during training; $L_1$ regularization is applied at every dense layer. In the case of CASTLE+, the causal graph used by CINN also serves as the causal DAG $\mathsf{G}^c$ specified by domain experts to be injected into CASTLE+. Injecting the same causal DAG enables a fair comparison of the two different means of incorporating causal DAG into neural network. Last but not the least, each benchmark method is initialized and seeded identically with the same random weights. 

In all the computational experiments, the entire dataset is randomly split into two parts: 80\% is used as the training set, while the remaining 20\% is reserved as the test set. Note that in the case of CINN, the training set is used to discover causal relationships and then train the neural network built upon the causal DAG after refinement. Then, ten-fold cross validation is used to examine the stability and robustness associated with the prediction performance of all the investigated models. At each fold, since each model converges at a different learning rate, a validation set (10\% of training data) is extracted from the training data to tune the model parameters, and those parameters resulting in the lowest mean squared errors (MSE) are treated as the ultimate set of parameters of the trained model. In doing so, the performance (e.g., learning rate, weight decay regularizer) of each model is tuned to its best state. Moreover, at each fold, we fix the seed of the program when splitting the entire data into training and test sets so that all the considered models have identical sets of data for training and test. Each model is trained using the Adam optimizer with the default learning rate of 0.001 for 800 epochs. Furthermore, an early stopping regime halts model training with a patience of 30 epochs.

In addition to comparing CINN with multiple state-of-the-art models, we also examine the performance of CINN with and without PCGrad to investigate the benefits of leveraging PCGrad in the optimization of neural network parameters. For a fair comparison, we also fix the seed value of the program when initializing the weights and biases of neural network so that both cases have the same set of initialized parameters as the starting point.

\subsection{Demonstration of CINN Setup}\label{subsec:cinn_setup}
\subsubsection{Causal Discovery}
As introduced earlier, CINN starts from discovering causal relationships from observational data. At the stage of causal discovery across different datasets, the values of $\lambda$ and $\tau$ in Eq.~(\ref{eq:no_tears}) need to be finetuned. For categorical features, one-hot-encoded representation is used for DAG discovery and CINN training, while the non-categorical features are standardized with mean 0 and variance 1. The FB dataset has 60 features in total after categorical features are encoded, because the month, weekday, and hour when a post is published are all categorical variables. In contrast, the BH dataset has only 14 features. Considering the significant variation associated with the number of features in each dataset, we set parameters $\lambda$ and $\tau$ to different values.

\begin{figure}[!ht]
    \centering
    \includegraphics[scale=0.5]{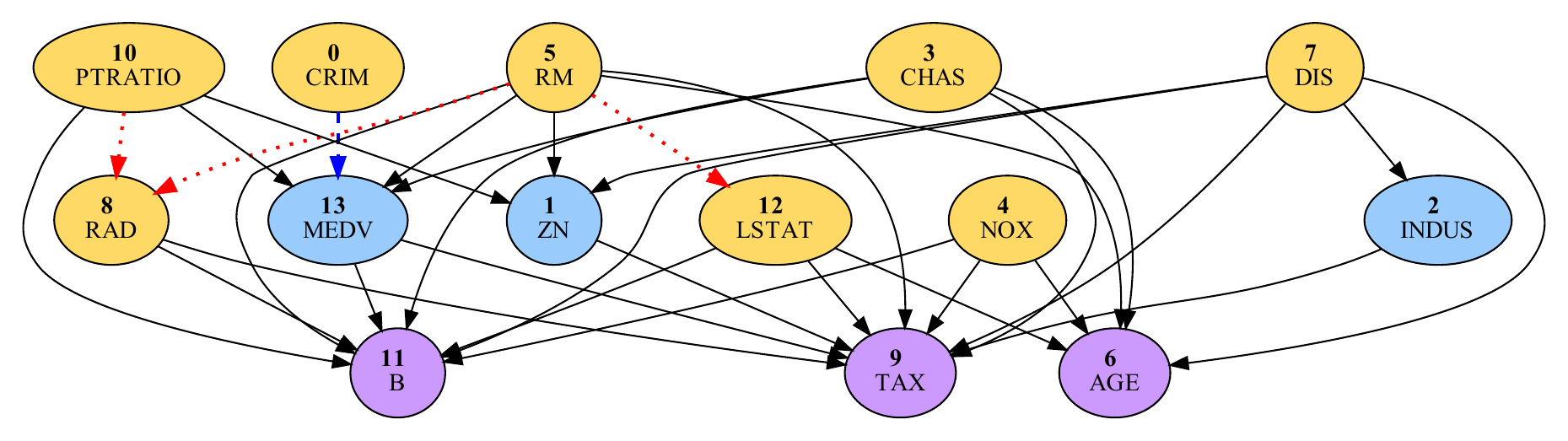}
    \caption{Discovery and refinement of causal relationships among observed variables for the BH dataset. The bold-faced number in each circle indicates the feature ID, while the associated text represents the feature name. Our primary task is to predict the MEDV value (node 13) using other relevant features. The circles in the same color belong to the same category of nodes. Specifically, circles in yellow refer to the set of root nodes $\mathsf{V}_C = \{ {\mathsf{X}_{0}}, {\mathsf{X}_{3}}, {\mathsf{X}_{4}}, {\mathsf{X}_{5}}, {\mathsf{X}_{7}}, {\mathsf{X}_{8}}, {\mathsf{X}_{10}}, {\mathsf{X}_{12}} \}$, circles in blue indicate the set of intermediate nodes $\mathsf{V}_B = \{ {\mathsf{X}_{1}}, {\mathsf{X}_{2}}, {\mathsf{X}_{13}} \}$, while circles in purple are the set of leaf nodes $\mathsf{V}_O = \{ {\mathsf{X}_{6}}, {\mathsf{X}_{9}}, {\mathsf{X}_{11}}\}$. The red dotted lines denote the edges eliminated from the discovered causal graph by exploiting expert knowledge, while the edge in blue denotes the link that is additionally added to the discovered causal graph.}
    \label{fig:boston_housing_cinn}
\end{figure}

\subsubsection{Causal DAG Refinement} After causal discovery, domain knowledge is then exploited to refine the discovered causal relationships for establishing CINN in two different ways: (i) refining the causal structure among observed variables by eliminating invalid causal links and adding substantiated causal edges; (ii) specifying and injecting quantitative relationships between certain variables into CINN following the method described in Section~\ref{subsec:cinn}. To demonstrate the specific steps of injecting expert knowledge into CINN construction, take the BH dataset as an example. Fig.~\ref{fig:boston_housing_cinn} illustrates the refined causal graph after expert knowledge is used to eliminate three links from the original causal graph discovered by the algorithm developed by~\cite{zheng2020learning}. Towards this goal, it is essential to verify if there is a causal relationship between one variable $a$ and another variable $b$ in a fast manner. A straightforward way to perform such verification is to imagine that we only perturb the value of variable $a$ while keeping the value of all the other variables fixed, and see whether the value of variable $b$ changes as a result of this perturbation~\citep{pearl2009causal}. Following this way, it is easy to observe that even if we change the value of attribute RM (average number of rooms, $\mathsf{X}_{5}$), it does not lead to a change on the values of attributes LSTAT (lower status of the population, $\mathsf{X}_{12}$), RAD (index of accessibility to radial highways, $\mathsf{X}_8$), among others. In contrast, if the value associated with the attribute CRIM (per capita crime rate by town, $\mathsf{X}_{0}$) is increased or decreased, it has a direct effect on the value of variable MEDV (median home value, $\mathsf{X}_{13}$). The causal relationship between CRIM ($\mathsf{X}_0$) and MEDV ($\mathsf{X}_{13}$) agrees with our common sense and, in fact, has been confirmed by several other causal discovery algorithms, such as conditional independence test~\citep{zhang2011kernel}, maximum likelihood estimator for causal discovery~\citep{wei2021nonlinear}, to name a few. As a result, we add the causal relationship between CRIM ($\mathsf{X}_{0}$) and MEDV ($\mathsf{X}_{13}$) in the discovered causal graph. It should be noted that the refined graph after elimination and/or addition operations must remain a DAG. If this is not the case, then necessary adjustments need be carried out to turn the refined graph into a DAG. 

In the refined causal graph shown in Fig.~\ref{fig:boston_housing_cinn}, the features in yellow circles are root nodes acting as input features in CINN, while the features in blue and purple circles are nodes in the intermediate and output layers, respectively. Given the refined causal DAG, we have $\mathsf{V}_{C} = \{ {\mathsf{X}_{0}}, {\mathsf{X}_{3}}, {\mathsf{X}_{4}}, {\mathsf{X}_{5}}, {\mathsf{X}_{7}}, {\mathsf{X}_{8}}, {\mathsf{X}_{10}}, {\mathsf{X}_{12}}\}$, $\mathsf{V}_{B} = \{ {\mathsf{X}_{1}}, {\mathsf{X}_{2}}, {\mathsf{X}_{13}}\}$, $\mathsf{V}_{O} = \{ {\mathsf{X}_{6}}, {\mathsf{X}_{9}}, {\mathsf{X}_{11}}\}$. When training CINN, the parameters are optimized so that the total loss over the intermediate nodes $\mathsf{V}_{B}$ and leaf nodes $\mathsf{V}_{O}$ is  minimized. In contrast, when making prediction, our primary task is to infer $\mathsf{X}_{13}$ (MEDV), which is a function of the input features $\mathsf{V}_{C}$. 

\subsubsection{Incorporation of Causal Relationships} In addition to eliminating and inserting causal links in the discovered causal graph, CINN is also equipped with the capability of incorporating quantitative relationship between two variables into the neural network. Again, take the BH dataset as an example. It is easy to understand that LSTAT ($\mathsf{X}_{12}$) has a negative impact on MEDV ($\mathsf{X}_{13}$), provided that all the other contributing factors (e.g., $\mathsf{X}_5$, $\mathsf{X}_3$, $\mathsf{X}_7$) remain fixed. In other words, the higher the LSTAT value, the lower the MEDV value. As a result, we can inject an inequality constraint into CINN, imposing that $\mathsf{X}_{13}$ and $\mathsf{X}_{12}$ are negatively causally related (i.e., ${\text{d}\mathsf{X}_{13}}/{\text{d}\mathsf{X}_{12}} \le  0$). To allow a margin of error, we further modify the constraint to ${\text{d}\mathsf{X}_{13}}/{\text{d}\mathsf{X}_{12}} - \varepsilon \le  0$, where $\varepsilon$ is set to 0.01. We can incorporate the quantitative relationship between MEDV ($\mathsf{X}_{13}$) and CRIM ($\mathsf{X}_{0}$) in a similar manner. It is worth highlighting that a significant difference between CINN and classical neural networks is that they treat all variables (except target variable $\mathsf{X}_{13}$) as inputs while the relationships among the input variables are ignored. Such an architecture design limits the possible ways of injecting causal knowledge, because it only permits to enforce relationships between target variable $\mathsf{X}_{13}$ and input variables. As shown in Table~\ref{tab:cinn_configuration}, CINN offers a much flexible way to incorporate causal knowledge into the neural network. Table~\ref{tab:cinn_configuration} gives a summary on the values of $\lambda$ and $\tau$, as well as causal DAG refinement and incorporation of causal relationships along the development of CINN for each dataset. More details can be found in the Appendix.

We adopt the same architecture as CASTLE~\citep{kyono2021towards} for all the alternative models that CINN is compared against (e.g., baseline, $L_1$, DO, BN). CINN has two hidden layers (Fig.~\ref{fig:cinn_architecture}) and ReLU is employed as the activation function. The first hidden layer has 32 units ($r=32$), and both the hidden layers $\theta_{21}, \theta_{22}, \cdots, \theta_{2q}$ ($q=16$) and $\theta_{31}, \theta_{32}, \cdots, \theta_{3u}$ ($u=16$) have 16 hidden units. The outputs of the layer $\theta_{2,1}, \theta_{2,2}, \cdots, \theta_{2,q}$ are then passed to the layers of intermediate nodes for further propagation. Finally, the outputs of the last intermediate layer and the outputs of the layer $\theta_{3,1}, \theta_{3,2}, \cdots, \theta_{3,u}$ are concatenated and mapped to the layer of output nodes via a hidden layer denoted by $\theta_{4,1}, \theta_{4, 2}, \cdots, \theta_{4,s}$, with $s=8$.

\subsection{Evaluation Metrics}
Because all the datasets considered are related to regression-type problems, we use MSE on the test set to quantify the performance of all the considered models with respect to the primary prediction task (the target variables are indicated in Table~\ref{tab:dataset_description}): 
\begin{equation}
{\rm{MSE}} = \frac{1}{N_{\rm{test}}}\sum\limits_{i = 1}^{N_{\rm{test}}} {{{\left( {{y_i} - {{\widehat y}_i}} \right)}^2}}, 
\end{equation}
where ${N_{\rm{test}}}$ denotes the number of samples in the test set, $y_i$ indicates the value of the target variable for the $i$-th sample, while ${\widehat y}_i$ is the prediction for the $i$-th sample made by the trained model.

\begin{table}[!ht]\small
    \centering
    \caption{Summary of causal DAG refinement and incorporation of causal relationships along the development of the CINN for each dataset}
    \begin{tblr}{Q[c, m, 2cm]Q[c, m, 1.2cm]Q[c, m, 1.2cm]Q[c, m, 3.2cm]Q[c, m, 2.2cm]Q[c, m, 4.2cm]}
    \hline
    \SetCell[r=2,c=1]{c}dataset & \SetCell[r=2,c=1]{c}\bm{$\lambda$} &  \SetCell[r=2,c=1]{c}\bm{$\tau$} &
     \SetCell[c=2]{c} Causal DAG Refinement
     &  & \SetCell[r=2,c=1]{c} Injection of Causal Relationships  \\
    \hline
     & & & \SetCell[r=1,c=1]{c} Elimination of discovered edges & \SetCell[r=1,c=1]{c} Injection of causal edges  & \\
    \hline
     BH    & 0.1 & 0.8 & [5, 8], [5, 12], [10, 8] & [0, 13] & $\frac{{\text{d}\mathsf{X}_{13}}}{{\text{d}\mathsf{X}_{12}}} - 0.01 \le  0$;
     $\frac{{\text{d}\mathsf{X}_{13}}}{{\text{d}\mathsf{X}_{0}}} - 0.01 \le  0$;
     $- \frac{{\text{d}\mathsf{X}_{13}}}{{\text{d}\mathsf{X}_{5}}} - 0.01 \le  0$;
     $\frac{{\text{d}\mathsf{X}_{11}}}{{\text{d}\mathsf{X}_{4}}} - 0.01 \le  0$;
     $\frac{{\text{d}\mathsf{X}_{11}}}{{\text{d}\mathsf{X}_{0}}} - 0.01 \le  0$.\\
     \hline
     WQ    & 0.1 & 0.8 & [11, 5], [11, 6], [1, 5], [10, 5], [0, 5], [3, 5],
                   [8, 3], [10, 3] & [5, 11], [6, 11], [3, 11], [4, 11], [7, 11], [10, 7]  & $\frac{{\text{d}\mathsf{X}_{7}}}{{\text{d}\mathsf{X}_{10}}} + 0.01 \le 0$ \\
     \hline
     FB    & 0.1 & 0.1 & [0, 6], [1, 2], [1, 6], [4, 3], [5, 1], [5, 2], [5, 6], [6, 2], [7, 1], [7, 2], [7, 6], [8, 1], [8, 2], [8, 6], [8, 7], [9, 0], [9, 1], [9, 2], [9, 5], [9, 6], [9, 7], [9, 8] & [1, 9], [2, 9], [5, 9], [6, 9], [7, 9], [8, 9], [41, 8] & None\\
     \hline
     BC    & 0.1 & 0.3 & [9, 0], [9, 4], [9, 8] & None & None\\
     \hline
     CM   & 0.01 & 0.05 & [0, 71], [7, 8], [8, 6], [12, 13], [13, 33], [17, 32], [20, 15], [20, 31], [21, 83], [28, 32], [31, 15], [31, 35], [31, 37], [40, 41], [41, 38], [43, 46], [43, 67], [44, 43], [44, 45], [44, 46], [45, 83], [48, 47], [56, 61], [58, 4], [60, 48], [61, 68], [64, 1], [65, 13], [67, 73], [67, 93], [68, 5], [71, 27], [71, 89], [71, 113], [79, 80], [80, 4], [80, 91], [81, 91], [91, 59], [93, 94], [94, 95], [95, 92] & [0, 27], [1, 66], [4, 80], [5, 68], [13, 12], [15, 122], [27, 71], [27, 89], [32, 28], [33, 13], [43, 122], [44, 122], [48, 122], [49, 122] & None \\
    \hline
    \end{tblr}
    \label{tab:cinn_configuration}
\end{table}


\subsection{Performance Comparison}\label{sec:performance_comparison}
After demonstrating the development process of CINN, we now compare its prediction performance against the other models under consideration. Note that the CINN model is trained for 800 epochs, and Adam optimizer with a learning rate of 0.001 is used to optimize the parameters. After CINN is trained, it can be used to make prediction for any variable in the intermediate and output layers. Since each dataset has a clearly defined target variable (see Table~\ref{tab:dataset_description}), we benchmark CINN against the other models with respect to predicting the target variables.

\begin{table}[!ht]
    \centering
    \caption{Performance comparison in terms of test MSE $\pm$ standard deviation using 10-fold cross-validation. Bold denotes the test MSE of CINN and underline indicates the test MSE of CASTLE and CASTLE+.}
    \begin{tabular}{lccccc}
    \toprule
      \diagbox[width=10em]{Model}{dataset}          & BH                & WQ                & FB & BC & CM \\
      \midrule
      Early stopping (Baseline)          & $0.117 \pm 0.036$ & $0.746 \pm 0.020$ & $0.162 \pm 0.030$ & $0.293 \pm 0.013$ & $0.407 \pm 0.027$\\
      $L_1$ norm              & $0.103 \pm 0.033$ & $0.736 \pm 0.020$ & $0.120 \pm 0.045$ & $0.284 \pm 0.016$ & $0.406 \pm 0.015$\\
      $L_2$ norm               & $0.111 \pm 0.038$ & $0.741 \pm 0.022$ & $0.115 \pm 0.042$ & $0.289 \pm 0.013$ & $0.396 \pm 0.029$\\
      Dropout 0.2       & $0.102 \pm 0.034$ & $0.732 \pm 0.011$ & $0.142 \pm 0.054$ & $0.286 \pm 0.006$ & $0.383 \pm 0.014$\\
      Batch Norm        & $0.130 \pm 0.035$ & $0.740 \pm 0.020$ & $0.282 \pm 0.042$ & $0.300 \pm 0.012$ & $0.435 \pm 0.010$\\
      Input Noise    & $0.109 \pm 0.040$  & 0.732 $ \pm 0.010$ & $0.120 \pm 0.036$ & $0.299 \pm 0.017$ & $0.395 \pm 0.019$\\
      MixUp    & $0.113 \pm 0.030$ & $0.754 \pm 0.023$& $0.159 \pm 0.042$ & $0.299 \pm 0.021$  & $0.425 \pm 0.034$\\
      CASTLE            & \underline{$0.101 \pm 0.019$} & \underline{$0.730 \pm 0.043$} & \underline{$0.114 \pm 0.098$} & \underline{$0.283 \pm 0.086$} & \underline{$0.380 \pm 0.032$}\\
      CASTLE+           & \underline{$0.093 \pm 0.015$} & \underline{$0.727 \pm 0.053$} & \underline{$0.112 \pm 0.064$} & \underline{$0.282 \pm 0.050$} & \underline{$0.368 \pm 0.022$}\\
      CINN (without PCGrad)  & \pmb{$0.083 \pm 0.006$} &\pmb{$0.710 \pm 0.020$}  & \pmb{$0.085 \pm 0.024$} &\pmb{$0.281 \pm 0.006$} & \pmb{$0.319 \pm 0.007$}\\
      CINN (with PCGrad)     & \pmb{$0.081 \pm 0.005$} & \pmb{$0.703 \pm 0.011$} & \pmb{$0.079 \pm 0.021$} & \pmb{$0.280 \pm 0.004$} & \pmb{$0.318 \pm 0.006$}\\
      \bottomrule
    \end{tabular}
    \label{tab:performance_comparison}
\end{table}

Table~\ref{tab:performance_comparison} summarizes the performance of these models across all the datasets under consideration. In the first place, we concentrate on comparing the general performance of CINN with other prevailing models in the literature. At a quick glance, CINN (with and without PCGrad) significantly outperforms other state-of-the-art models in terms of test MSE across most datasets; an exception is BC, for which CINN leads to a slight improvement in prediction performance compared to other alternative models. It is worth highlighting that CASTLE only leads to a marginal improvement in prediction performance compared with other prevailing regularizers, while CINN outperforms CASTLE significantly, as reflected by the substantial reduction in the test MSE. For example, with respect to the BH dataset, CASTLE slightly beats the best-performed regularizer (i.e., DO); however, compared with CASTLE, CINN reduces the test MSE by 17.8\% and 20\%, respectively. Moreover, CINN diminishes the test MSE substantially in a consistent and stable manner as observed in other datasets, such as WQ, FB, and BC. In particular, in the case of CM, CINN reduces the test MSE from 0.383 to 0.319 (without PCGrad) and 0.318 (with PCGrad), which is a nearly 18\% reduction against CASTLE. An interesting observation is that CASTLE+ achieves superior performance to CASTLE due to the incorporation of causal DAG, which underscores the importance of encoding causal DAG into neural networks. 


Another appealing feature of CINN is its capability of attaining a desirable stability in the prediction performance. As shown in Table~\ref{tab:performance_comparison}, CINN reduces the standard deviation of test MSE to a large extent when benchmarking against other alternative models. The high variability associated with these alternative models is attributed to the inclusion of spurious correlations and relationships when building the deep learning models. In contrast, structuring the architecture of CINN as informed and guided by the hierarchical causal relationships extracted from observational data and domain experts substantially enhances the stability of prediction performance. In fact, the reduction in the standard deviation of test MSE is much more salient than the corresponding reduction in the associated mean value. The elevated stability in prediction performance is essential to the adoption of neural networks in safety-critical applications.

Next, we focus on performance comparison within the CINN family. As mentioned earlier in Section~\ref{sec:pcgrad}, CINN involves the learning of multiple tasks that are conflicting with respect to gradients in nature. Since the gradients associated with these learning tasks differ significantly in magnitude and direction, PCGrad is leveraged to mitigate gradient interferences in multi-task learning by eliminating conflicting components from the set of gradients. Computational results reported in Table~\ref{tab:performance_comparison} suggest that integrating PCGrad into CINN not only leads to a lower mean MSE, but also results in a slightly contracted standard deviation when compared with the case of no PCGrad. These findings reveal the efficacy of utilizing PCGrad to deconflict the gradients when training the CINN model. In practice, PCGrad is recommended to be used in the context of multi-task learning to enhance model performance in prediction and stability.

\subsection{Robustness Checks}\label{sec:robustness_checks}
The aim of this subsection is to carry out a series of robustness checks for the developed CINN. Specifically, we investigate the effect of two factors on the CINN's performance. The first is the seed value for initialization of neural network parameters (e.g., bias, weights), and the second is regarding two hyperparameters (i.e., $\gamma$ in Eq.~(\ref{eq:overall_loss}) and learning rate $\beta $).

\begin{figure}[!ht]
    \centering
    \includegraphics[scale = 0.39]{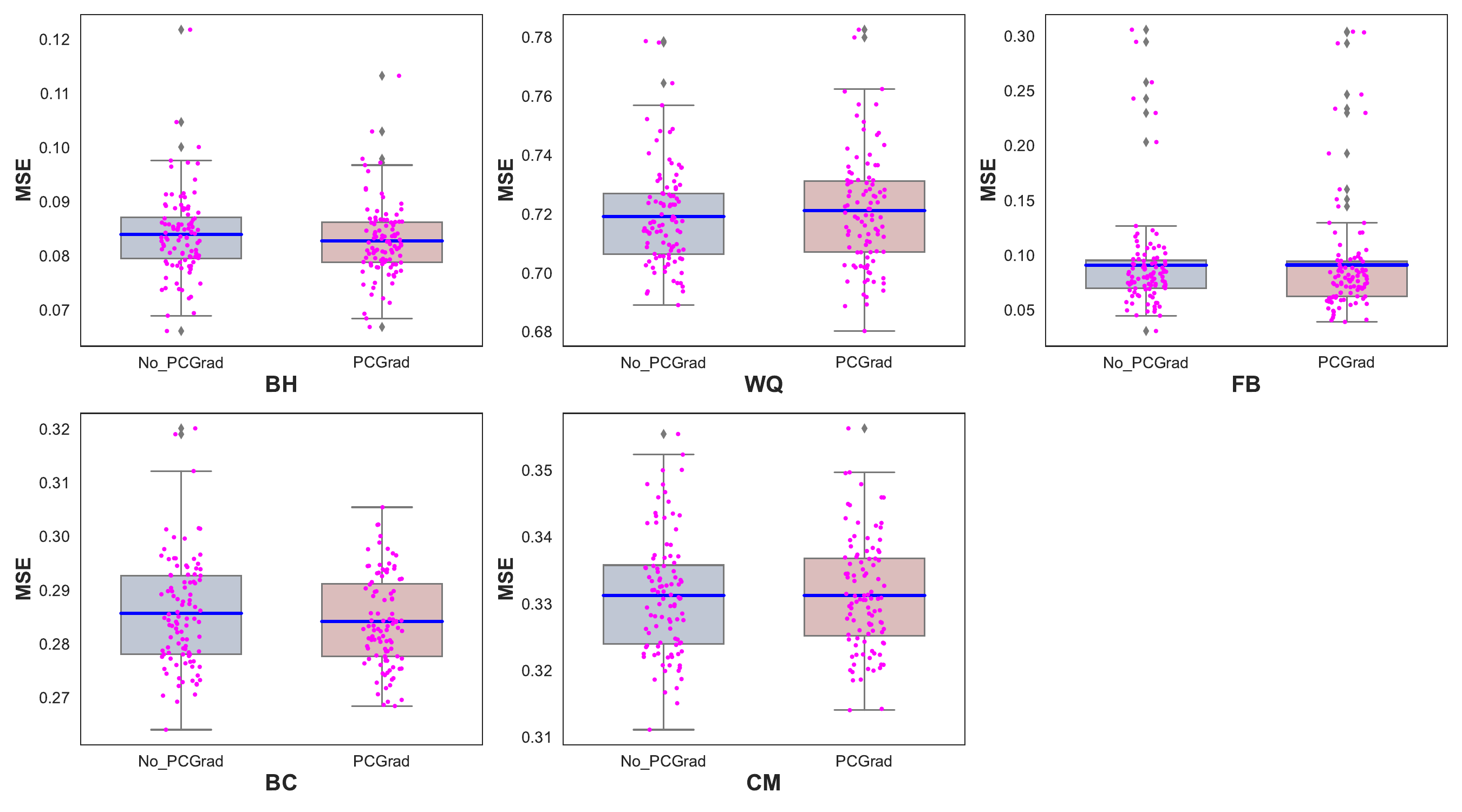}
    \caption{Effect of seed value on the robustness of CINN performance. The solid blue line indicates the mean value of MSE. Note that the 100 MSE values (10 seed values $\times$ ten-fold cross validation) are aggregated in the box plot. The low deviation of MSE value reflects the performance robustness of CINN with respect to the seed value.}
    \label{fig:robustness_check}
\end{figure}

For the seed value, we use ten different seed values to initialize the CINN parameters. In each fold of cross validation, we run 10 experiments with an identical seed value for initialization. The other CINN configurations (e.g., learning rate, number of epochs, network architecture) remain the same as before. Fig.~\ref{fig:robustness_check} shows the relationship between the seed value and the CINN performance. Clearly, as reflected by the variation of MSE value, the effect of different seed values on the CINN's prediction performance is trivial, and CINN exhibits a relatively robust performance that is consistent with the computational results reported in Table~\ref{tab:performance_comparison}. As shown in Fig.~\ref{fig:robustness_check}, CINN achieves a much lower mean MSE value than the other baseline models across all datasets. Furthermore, when compared with the case without PCGrad, CINN with PCGrad achieves a lower variance for datasets BH, WQ, and CM, while maintaining a comparable performance for FB and BC. 

\begin{figure}[!ht]
    \centering
    \includegraphics[scale = 0.38]{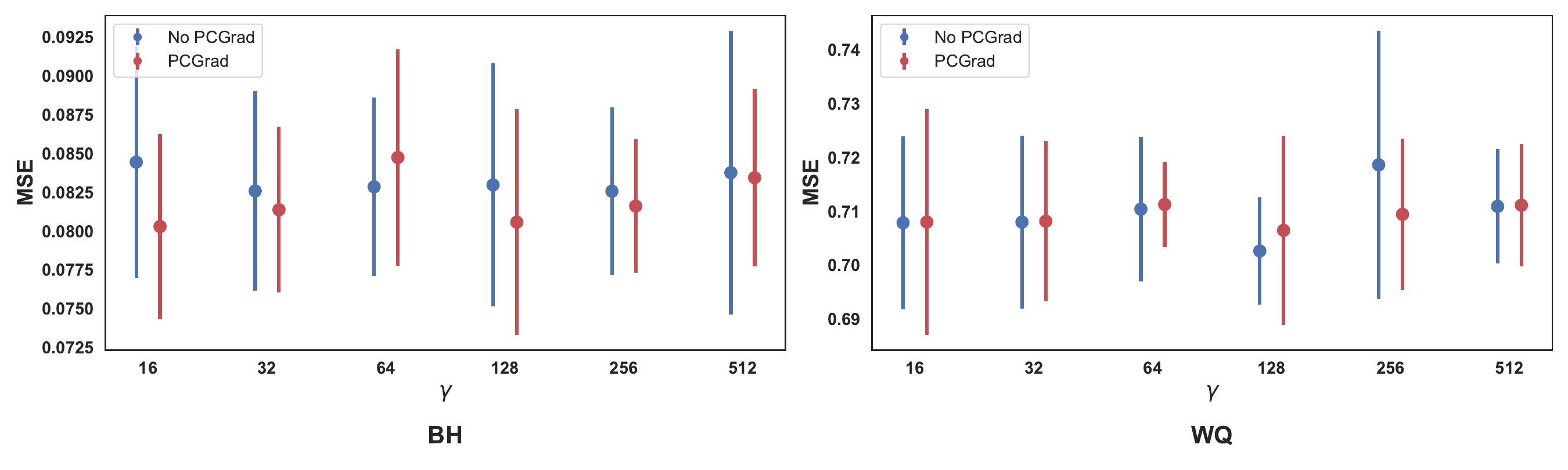}
    \caption{Impact of hyperparameter $\gamma$ on the robustness of CINN performance.}
    \label{fig:CINN_weight_effect}
\end{figure}

We then investigate the impact of hyperparameter $\gamma$ on the performance of CINN. Recall that $\gamma$ represents the weight associated with the domain prior on the causal relationship in Eq.~(\ref{eq:overall_loss}). As $\gamma$ is relevant only when domain prior is involved in the construction of CINN, we confine the robustness check to datasets BH and WQ, as no domain prior is imposed with respect to FB, BC, and CM (see the last column in Table~\ref{tab:cinn_configuration}). Fig.~\ref{fig:CINN_weight_effect} illustrates the impact of $\gamma$ on the robustness of the CINN's performance. At a quick glance, the effect of $\gamma$ on the performance of CINN is insignificant, as reflected by the negligible variation in the MSE values for the three datasets. Notably, regardless of the value of $\gamma$, the mean MSE values stay consistent with those reported in Table~\ref{tab:performance_comparison}. The insignificant effect of $\gamma$ on the CINN's performance is primarily attributed to the small magnitude of loss $\mathcal{L}_R$ compared to the overall loss pertaining to the observed variables in the intermediate and output layers.\footnote{Note that $\mathcal{L}_R$ is not always active and it becomes active only when the causal relationship described in the last column of Table~\ref{tab:cinn_configuration} is violated.}

\begin{figure}[!ht]
    \centering
    \includegraphics[scale = 0.38]{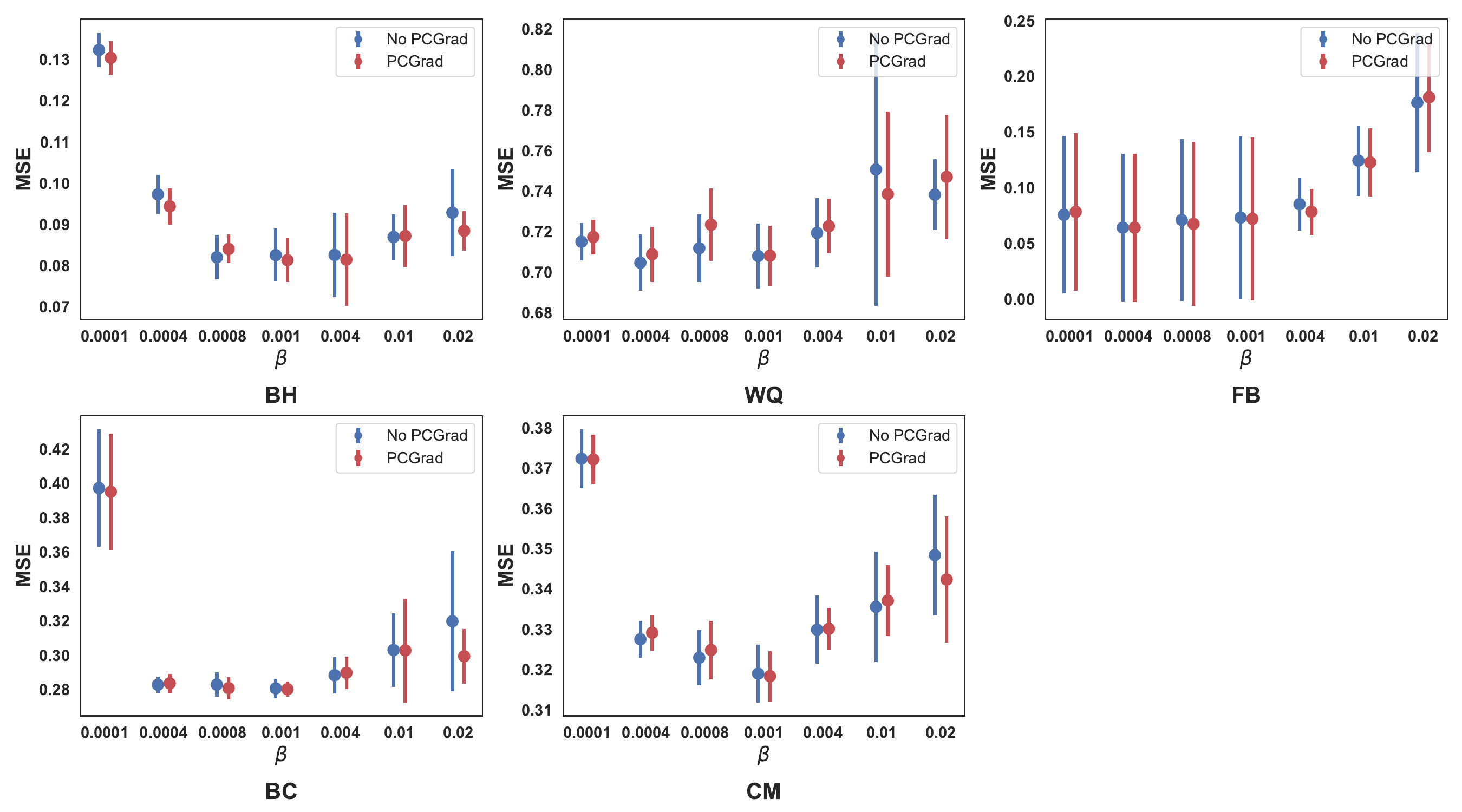}
    \caption{Impact of learning rate $\beta$ on the robustness of CINN performance}
    \label{fig:CINN_lr_effect}
\end{figure}

We further study the robustness of CINN with respect to the learning rate $\beta$. For this purpose, we gradually increase the value of $\beta$ from 0.0001 to 0.02 and observe the change in the CINN's performance. As shown in Fig.~\ref{fig:CINN_lr_effect}, the learning rate has a strong effect on the performance of CINN. In particular, when $\beta$ takes a large value (say, 0.02) or an extremely low value (say, 0.0001), the performance of CINN deteriorates substantially across all the five datasets. For example, with respect to the CM dataset, the MSE value increases to 0.37 when $\beta$ drops down to 0.0001. While for other values of $\beta$, the MSE value fluctuates within a reasonable range. Another interesting observation is that CINN with PCGrad outperforms the case without PCGrad in either the MSE value or the standard deviation\textemdash reflected by the spread of the bar.

\subsection{Ablation Study of CINN}
Intuitively, aligning a neural network with an accurate and informative causal structure should yield an improved prediction performance. The aim of this subsection is to examine the effect of (i) leveraging expert knowledge in rectifying the discovered causal graph and (ii) enforcing quantitative causal relationships in the development of CINN, in an incremental manner. We are particularly interested in whether and how much the prediction performance of CINN can be improved if the neural network is enforced to abide by ``guidance" provided by domain experts. Let us consider datasets BH and WQ for demonstration purposes. The ablation study of CINN with respect to the other three datasets can be found in the Appendix. We perform a series of experiments by incrementally injecting expert knowledge in the form of causal links and quantitative causal relationships into CINN so as to examine the effect of incorporating these information on the performance of the primary prediction task. In this ablation study, the specific expert knowledge incorporated into CINN at each step for BH and WQ is summarized in Table~\ref{tab:ablation_study}. Specifically, we consider 7 different sets of expert knowledge for both datasets. The steps shaded in gray are expert knowledge in the form of quantitative causal relationships, while those in blue are expert knowledge in the form of trimming the discovered causal graph in structure. 

\begin{table}[!ht]\small
    \centering
    \caption{Expert knowledge injected into the construction of CINN at each step for BH and WQ}
    \label{tab:ablation_study}
    \begin{tabular}{L{2.5em}L{22em}L{22em}}
    \toprule
    \textbf{Step} &  \textbf{BH}   &  \textbf{WQ} \\
    \hline
    \textbf{1} & Inject no expert knowledge & \cellcolor{blue!15}Reverse the direction of link [5, 11] to build CINN with the discovered causal graph because node 11 is the target variable in WQ \\
    \hline
    
    \textbf{2} & \cellcolor{blue!15}Eliminate edges [5, 8], [10, 8], [5, 12] from the discovered causal graph & \cellcolor{blue!15}Reverse the direction of link [6, 11]\\
    \hline
    
    \textbf{3} & \cellcolor{blue!15}Insert edge [0, 13] into the discovered causal graph &  \cellcolor{blue!15}Remove [1, 5], [10, 5], [0, 5], [3, 5]  \\
    \hline

    \textbf{4} & \cellcolor{gray!25}(MEDV w.r.t. LSTAT) -- Impose the causal relationship between $\mathsf{X}_{13}$ and $\mathsf{X}_{12}$ as $\frac{{\text{d}\mathsf{X}_{13}}}{{\text{d}\mathsf{X}_{12}}} - 0.01 \le 0$ in CINN & \cellcolor{blue!15}Remove [3, 6]\\
    \hline
    
    \textbf{5} & \cellcolor{gray!25}(MEDV w.r.t. RM) -- Inject the causal relationship between $\mathsf{X}_{13}$ and $\mathsf{X}_{5}$ as $0.01 - \frac{{\text{d}\mathsf{X}_{13}}}{{\text{d}\mathsf{X}_{5}}}  \le 0$ in CINN & \cellcolor{blue!15}Remove [8, 3], [10, 3] and add [4, 11], [7, 11]\\
    \hline
    
    \textbf{6} & \cellcolor{gray!25}(MEDV w.r.t. CRIM) -- Impose the causal relationship between $\mathsf{X}_{13}$ and $\mathsf{X}_{0}$ as $\frac{{\text{d}\mathsf{X}_{13}}}{{\text{d}\mathsf{X}_{0}}} - 0.01 \le 0$ in CINN & \cellcolor{blue!15}Add [3, 11]\\
    \hline
    
    \textbf{7} & \cellcolor{gray!25}(B w.r.t. CRIM and NOX) -- Inject the causal relationships between $\mathsf{X}_{11}$ and $\mathsf{X}_{0}$, $\mathsf{X}_{4}$ as $\frac{{\text{d}\mathsf{X}_{11}}}{{\text{d}\mathsf{X}_{0}}} - 0.01 \le 0$ and $\frac{{\text{d}\mathsf{X}_{11}}}{{\text{d}\mathsf{X}_{4}}} - 0.01 \le 0$ in CINN & \cellcolor{gray!25}(total sulfur dioxide w.r.t. alcohol) -- Enforce the quantitative causal relationship between $\mathsf{X}_6$ and $\mathsf{X}_{10}$ as $\frac{{\text{d}\mathsf{X}_{6}}}{{\text{d}\mathsf{X}_{10}}} + 0.01 \le 0$ \\
    \bottomrule     
    \end{tabular}
\end{table}

\begin{figure}[!ht]
    \subfigure[BH]{\includegraphics[width=0.49\textwidth]{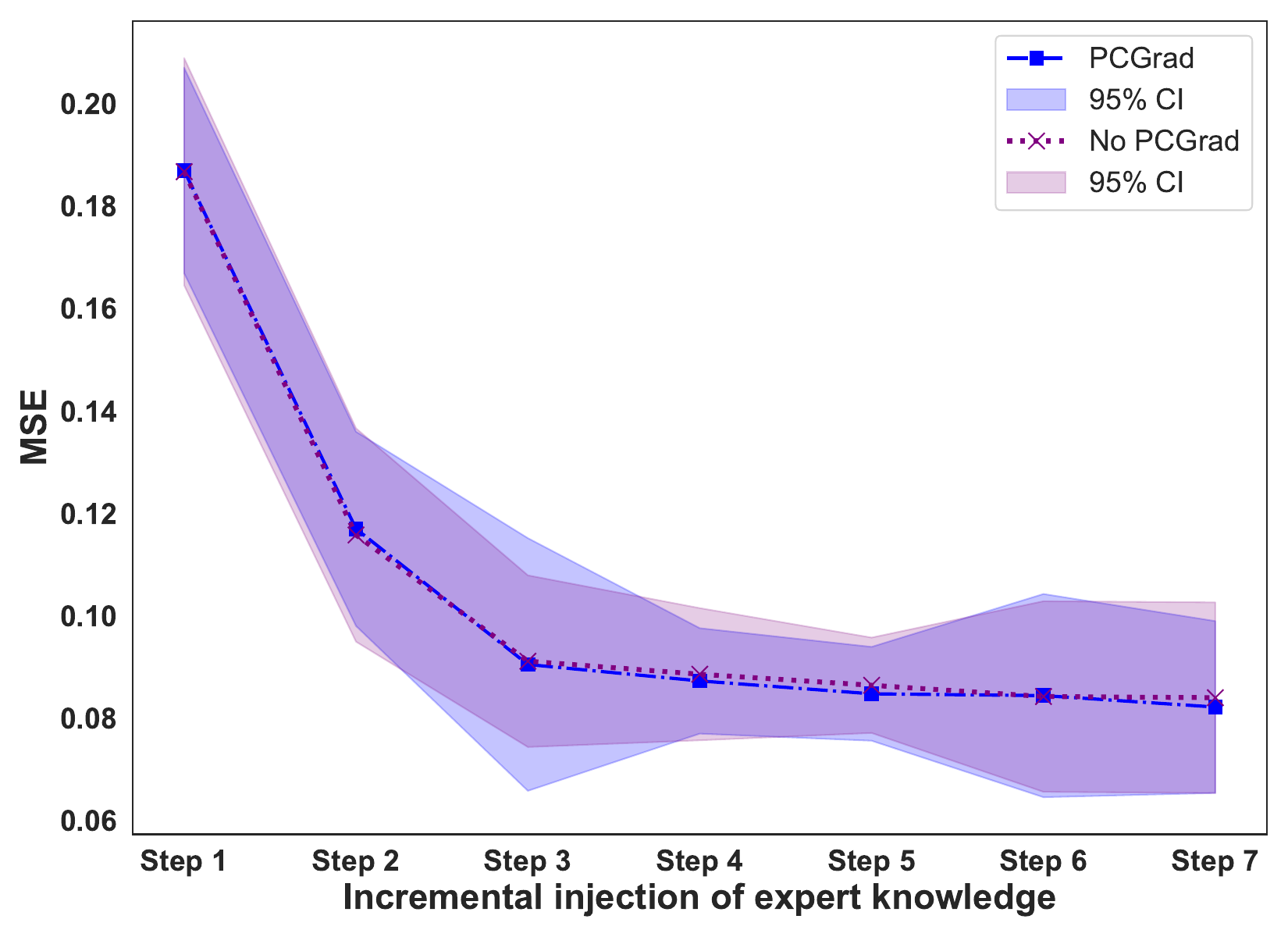}}
    \subfigure[WQ]{\includegraphics[width=0.49\textwidth]{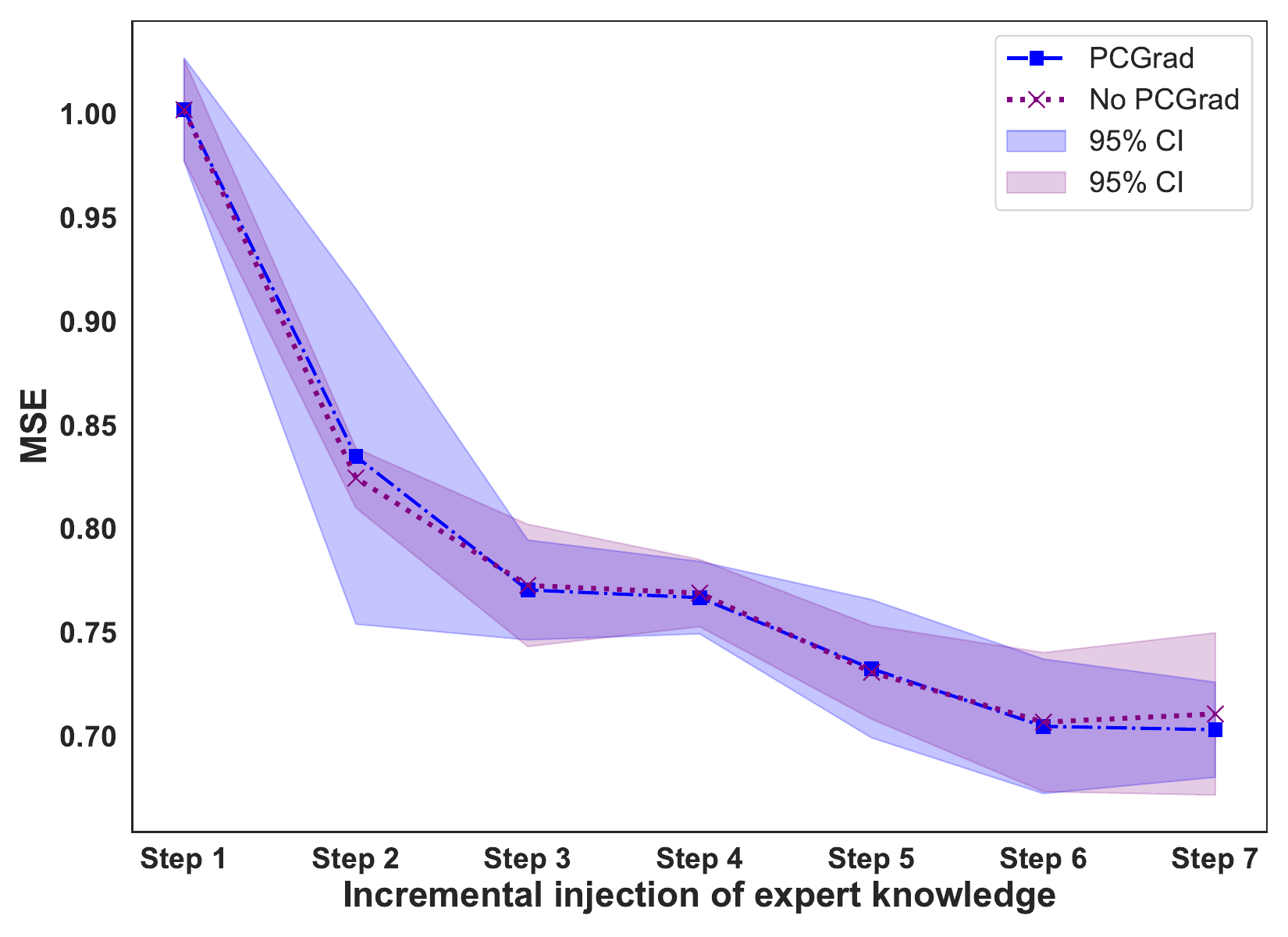}}
    \caption{Effect of incrementally injecting expert knowledge on the test MSE for (a) BH and (b) WQ}
    \label{fig:bh_ablation_study}
\end{figure}

\begin{figure}[!ht]
    \subfigure[BH at Step 1]{\includegraphics[width=0.48\textwidth]{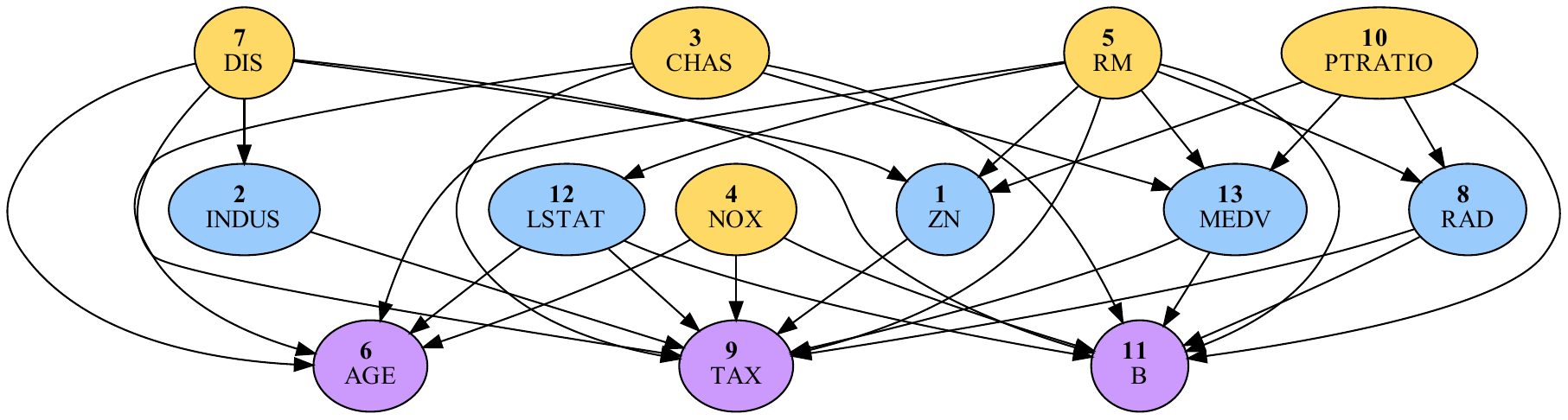}}
    \subfigure[BH at Step 2]{\includegraphics[width=0.48\textwidth]{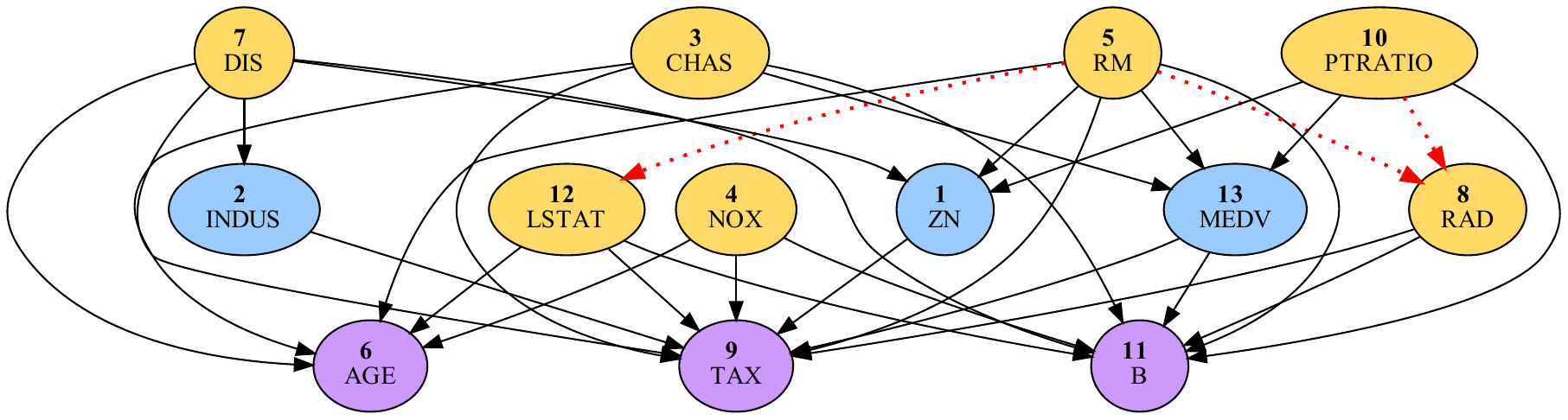}}

    \subfigure[WQ at Step 1]{\includegraphics[width=0.48\textwidth]{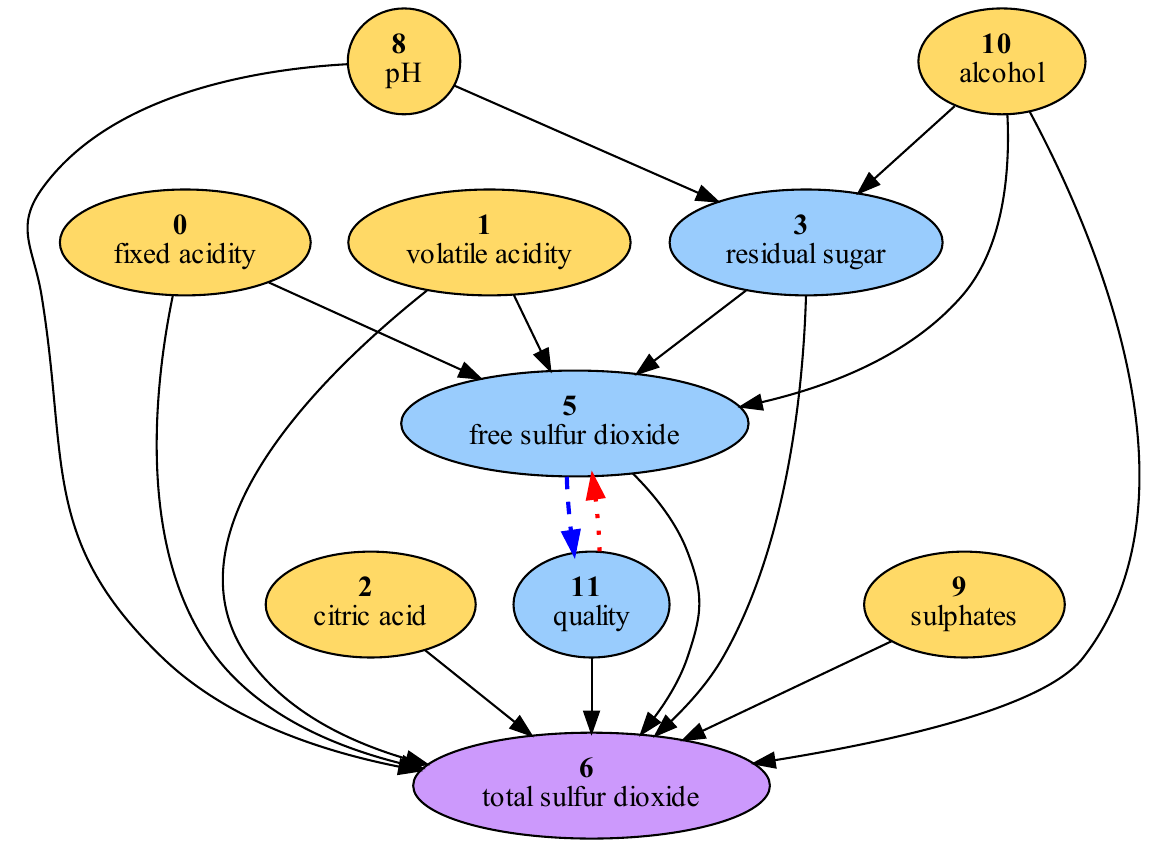}}
    \subfigure[WQ at Step 2]{\includegraphics[width=0.48\textwidth]{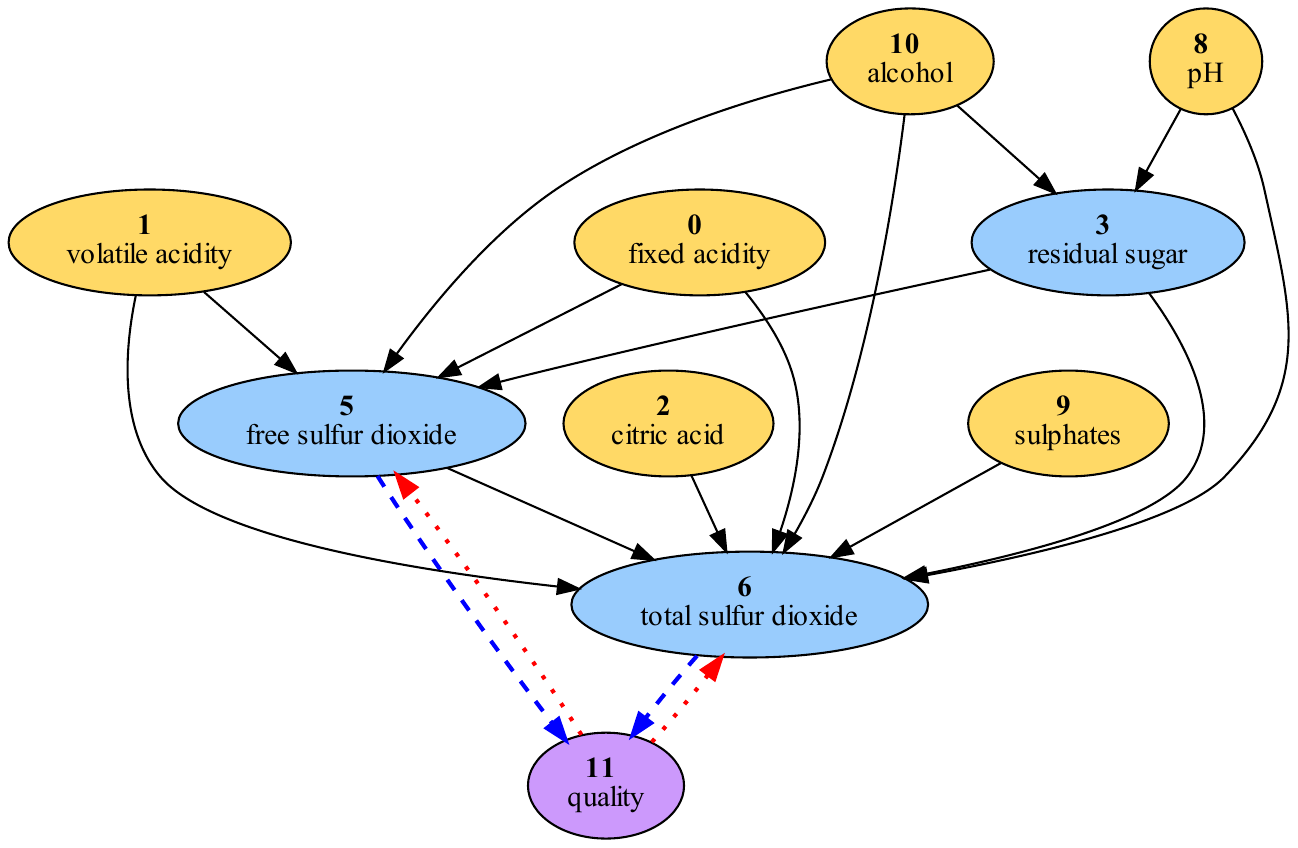}}
    \caption{Evolution of causal graph. 
    (a) BH Step 1: $\mathsf{V}_C = \left\{3, 4, 5, 7, 10 \right\}$, $\mathsf{V}_O = \left\{6, 9, 11 \right\}$, $\mathsf{V}_B = \left\{1, 2, 8, 12, 13 \right\}$;
    (b) BH Step 2: $\mathsf{V}_C = \left\{3, 4, 5, 7, 8, 10, 12 \right\}$, $\mathsf{V}_O = \left\{6, 9, 11 \right\}$, $\mathsf{V}_B = \left\{1, 2, 13\right\}$;
    (c) WQ Step 1: $\mathsf{V}_C = \left\{0, 1, 2, 8, 9, 10 \right\}$, $\mathsf{V}_O = \left\{6 \right\}$, $\mathsf{V}_B = \left\{[3], [5], [11] \right\}$;
    (d) WQ Step 2: $\mathsf{V}_C = \left\{0, 1, 2, 8, 9, 10 \right\}$, $\mathsf{V}_O = \left\{11 \right\}$, $\mathsf{V}_B = \left\{[3], [5], [6]\right\}$;
    }
    \label{fig:cinn_dissect}
\end{figure}
    
In the ablation study, ten-fold cross validation is used to examine the performance of CINN with respect to the primary prediction task (i.e., inferring the median value of owner-occupied homes (i.e., MEDV) for BH and wine quality for WQ, respectively). Fig.~\ref{fig:bh_ablation_study} shows the effect of incrementally incorporating expert knowledge on the prediction performance of CINN. Consistent with our expectation, the test MSE drops down in a steady manner when more expert knowledge is utilized to build CINN. In particular, if no expert knowledge is exploited (Step 1), then CINN has an average test MSE of 0.187 for dataset BH, which is much larger than that of CASTLE. In case of BH, after incorporating these quantitative causal relationships, the mean MSE of PCGrad drops from 0.090553 (Step 3) to 0.087347 (Step 4), 0.084830 (Step 5), 0.084479 (Step 6), and 0.082246 (Step 7). 

We have similar finding for WQ\footnote{As node 11 is the quantity of interest to be predicted by the model, if the direction of the link associated with node 11 is not reversed in the discovered causal graph, then node 11 becomes a root node and the CINN model cannot function properly as it is unable to predict the value of node 11. This is why link $[5, 11]$ is reversed in the Step 1 of WQ.}. After quantitative causal relationship is injected at Step 7, the mean MSE of PCGrad drops from 0.7049 (Step 6) to 0.7032 (Step 7). In addition, as reflected by the slope of the curve in Fig.~\ref{fig:bh_ablation_study}, the reduction in test MSE is most significant from Step 1 to Step 2 in both cases, as the change in the causal graph structure is most informative. Fig.~\ref{fig:cinn_dissect} illustrates the evolution of causal graph from Step 1 to Step 2. In the case of BH, after removing three invalid causal links, the nodes in the only intermediate layer is reduced from 5 to 3 (LSTAT and RAD become input nodes rather than intermediate nodes after refinement). While in the case of WQ, since total sulfur dioxide is also a significant factor affecting wine quality, after reversing the direction of link $[6, 11]$ the test MSE decreases significantly.

The ablation study highlights that the difference in the test MSE values between Step 1 and Step 7 is attributed to the injection of informative causal knowledge in the form of structural and functional causal relationships. In essence, effectively integrating structural and functional causal knowledge drives the learning of relationships among observed variables by the neural network.

\section{Discussion}\label{sec:discussion}
\subsection{Implications for Research}
The proposed CINN methodology yields several important implications for research in ML. First, when developing deep neural networks, most prior studies have either ignored the causal relationships in observational data~\citep{craja2020deep} or implicitly injected causal DAG through a customized loss function~\citep{kyono2020castle}. For those considering causal relationships, they often neglect the orientation associated with the causal relationships~\citep{kyono2020castle,russo2022causal}. In this paper, we develop a generic CINN by explicitly encoding the hierarchical causal DAG discovered from observational data into the design of neural network architecture. In doing so, the causal DAG serves as an informative prior to guide the architecture design while those not in adherence to the structure of causal DAG are eliminated. As a side product, establishing an explicit correspondence between nodes in the causal DAG and units in the neural network facilitates the incorporation of domain knowledge on stable causal relationships, no matter whether the causal relationship is expressed in a qualitative or quantitative form.

Second, the designed loss function for CINN represents a significant departure from conventional ones, as most deep learning models are trained to minimize the prediction error with respect to a single target variable~\citep{yang2023getting}. With the explicit mapping of hierarchical causal DAG into CINN, the associated loss function is devised to drive co-learning of causal relationships among different classes of nodes (say, $\mathsf{V}_C$ and $\mathsf{V}_B$, $\mathsf{V}_B$ and $\mathsf{V}_O$) by minimizing the total loss over the nodes in both the intermediate and output layers. Formulating such a learning process increases the CINN's robustness, as each mapped outcome variable in the succeeding layer serves as an informative target output (or constraints) on the predictors in the preceding layer. From an optimization standpoint, injecting these constraints substantially improves the CINN's stability and robustness (see Sections~\ref{sec:performance_comparison} and~\ref{sec:robustness_checks}), because the feasible regions of neural network parameters are pruned and any parameter combination violating these constraints would be discarded immediately.

Third, while most existing studies treat neural network training as a standalone task with nearly no or limited input from humans~\citep{wu2021fairplay}, the developed CINN offers a straightforward interface for incorporating expert knowledge by two principal means. This highlights another salient difference between CINN and existing methods\textemdash that is, CINN keeps humans in the loop: end users can easily communicate background information and business context with the neural network to shape the way the algorithm ``thinks” via a shared language of causality. The value of synergizing context-specific and theory-driven knowledge with deep learning models has also been demonstrated in the information systems literature~\citep{chen2023theory}. Specifically,~\cite{chen2023theory} combined personalized product preferences with a theory-driven modeling of customer dynamic satisfaction and achieved superior performance in customer response prediction. In this study, CINN allows experts to augment neural networks with knowledge on the cause-and-effect relationships that underlie observational data, while traditional deep learning models normally have no access to such knowledge. In the context of CINN, domain knowledge can be exploited to either refine the causal DAG discovered from observational data (by removing invalid causal edges and/or adding substantiated causal relationships overlooked by the causal discovery algorithm), or impose quantitative or qualitative constraints on stable causal relationships to prevent deep learning models from making misleading predictions. The human-machine symbiosis/collaboration complements the strength of each other, facilitating the realization of trustworthy ML~\citep{choudhury2020machine}.

\subsection{Implications for Practice}
Our findings generate broad implications for practitioners adopting deep learning models in operational research and ML communities. First, our work underscores the importance of causal DAG knowledge in strengthening the prediction performance of neural networks via thorough computational experiments and ablation study. The developed CINN offers a better way to mine, organize, and utilize information embodied in observational data. Considering the significant value of causal DAG, the data that have been used to train deep learning models in the past can be fully leveraged in the following two steps: discovering causal DAG and getting such DAG injected into neural network subsequently. To this end, the current pipeline for building deep learning models can be readily modified to reinforce the extant model as CINN. Specifically, rather than building a deep learning model with observational data directly, the value of observational data can be exploited twice: causal DAG discovery for the first time, and training the deep learning model for the second time. With such a minor modification, the value of observational data is fully exploited to enhance the prediction performance of neural networks.


Second, our ablation study reveals that domain knowledge also plays an essential role in enhancing the robustness and stability of neural networks. Such findings have important implications for practice. In particular, many industries (e.g., supply chain, finance, e-commerce, manufacturing) have already accumulated abundant domain knowledge over time that can help to understand system behaviors in various scenarios. While most data-driven models discard away valuable information in domain knowledge, we argue that the knowledge from multiple sources (e.g., system flowcharts, simulation study, system modeling, historical data, domain experts) should be fused together to augment the causal DAG discovered from observational data. Knowledge-driven learning offers a seamless synthesis of both paradigms: powerful representation learning capability of deep neural network, and alignment of fundamental mechanics during modeling. To this end, practitioners can elicit opinions from experts and incorporate such unique domain knowledge into the learning process. Taking manufacturing lines as an example, the process engineer might know that temperature in sensor 1 has a linear positive relationship with pressure in chamber X. With CINN, practitioners are able to embed such kind of knowledge into the neural network and ensure that the neural network always respects this relationship. In this sense, CINN combines the powers of both domain expertise and data-driven approaches.

These attractive features in CINN have profound implications to a wide range of practitioners, such as retail operations manager, supply chain manager, market analyst, manufacturing industry, to name a few. For example, in the retail industry, by relying on the causal drivers of purchasing behavior to build the CINN model, retail operations manager could gain an accurate and generalized forecast on product sales and then optimize subsequent retail operations (e.g., stock level) accordingly; in the field of supply chain, by combining the expertise of supply chain managers with the power of algorithmic causal discovery, a CINN model can be developed to forecast order delay and the likelihood of delays during the order life cycle; supply chain managers could utilize the developed model to understand the propagation of delay throughout the supply chain, identify the bottlenecks across the order life cycle, and perform ``what-if" analysis to seek recommendations in mitigating order delay. In the market campaign, by characterizing the causal drivers of campaign performance and the relationships among these factors in neural network, market analyst could use the resultant CINN model to attribute the causal effects of customer conversion, perform ``what-if" analysis to simulate how the market campaign performance would differ in varying scenarios, and optimize the allocation of marketing budget.


\subsection{Limitations and Future Research}
Our research presents some limitations that deserve future research efforts. First,  in this study expert knowledge is injected after causal DAG discovery. It is interesting to investigate another alternative that imposes expert knowledge at the DAG discovery stage, where expert knowledge can be represented as hard constraints in the optimization problem such that the discovered DAG strictly conforms to well established human knowledge. This improves the automation of CINN development. To this end, the causal discovery algorithm used in this paper needs to be modified accordingly, and it is particularly important to investigate\textemdash in theoretical and experimental manners\textemdash whether the human-guided causal discovery algorithm still converges to a stable solution after introducing such constraints.

Second, it is meaningful to integrate DAG discovery and CINN construction to form an end-to-end CINN optimization and learning pipeline. In particular, with respect to optimization, the hierarchical modeling of causal relationships mapped into the neural network poses challenges for classical gradient descent optimizers. This is because highly coupled relationships arise among multiple learning tasks, which makes the optimization space of neural network parameters extremely intricate. 
It is thus necessary to develop dedicated optimization algorithms for CINN by explicitly considering the coupled relationships among the multiple learning tasks. 

Thirdly, causal structure discovery plays a pivotal role in the construction of CINN. As causal discovery typically comes with strong assumptions (e.g., causal sufficiency, faithfulness, linear vs nonlinear causal relationship), errors in the discovered causal structure could propagate to the downstream tasks in the CINN development and spoil the CINN's performance ultimately. It is therefore important to assess the impact of deviation in the identified causal structure on the CINN's performance and develop effective measures to mitigate such effect.  

Finally, the transparent mapping of hierarchical causal DAG into a neural network together with strict preservation of the orientation of causal relationships opens the door for performing intervention queries (referred to as $do$ operator in causal graph) with CINN. As described in the structural causal framework by~\cite{pearl2000models}, the effect of an intervention on a given variable $\mathsf{X}_i$ has on another variable $\mathsf{X}_j$ can be modelled with the \textit{do} operator. For instance, $\Pr({\mathsf{X}_j}|do(\mathsf{X}_i = a))$ measures the effect of an intervention action on $\mathsf{X}_i$ has on the variable $\mathsf{X}_j$. The \textit{do} operation can be translated into an equivalent mutilation of graph topology of the causal DAG \citep{pearl2000models}\textemdash that is, we simply remove all links inbound to the node $\mathsf{X}_i$ from the network, and assign $\mathsf{X}_i$ the value of $a$. In a similar way, the mutilation of the causal DAG can be translated into CINN by removing corresponding links and assigning the intervened variable the fixed value. By doing so, CINN moves beyond prediction to answer interventional queries that traditional deep learning models are unable to do. Accurate estimation of causal effect with CINN has broad implications to operational research across many domains ranging from application spillover effect estimation to purchase behavior analysis~\citep{tripathi2022analyzing}. In the future, it is highly valuable to explore intervention queries with CINN and evaluate the performance of CINN in intervention queries compared to other alternatives in the literature. 

\section{Conclusion}\label{sec:conclusion}
In this paper, we develop a generic interface for encoding structural and relational causal knowledge among observed variables into neural network to guide its learning. In concert with the causal DAG-informed neural network architecture, CINN is devised to minimize the total loss over the variables in the intermediate and output layers to drive co-learning of causal relationships among observed variables. By exploiting the causal DAG as an inductive bias to the learning of neural network, CINN exhibits a superior performance than other state-of-the-art alternatives. Thhrough robustness checks and ablation study, we demonstrate that causal knowledge contributes to a significantly enhanced performance of neural network. The proposed research deepens our understanding on the nuanced role of causal knowledge in enhancing the neural network performance and highlights the importance of capturing underlying variable relationships when developing deep learning models. Importantly, the devised generic interface moves neural networks beyond pure data-driven learning paradigm by permitting to incorporate domain expertise in various forms in a flexible manner. Integrating domain knowledge into CINN complements the shortcoming of pure data-driven learning in the lack of contexts, thus leading to an augmentation in the prediction performance.

\bibliographystyle{apa}
\biboptions{authoryear}
\bibliography{ref}

\end{document}